\documentclass[letterpaper,twocolumn,10pt]{article}
\usepackage{usenix,epsfig,endnotes}

\usepackage{amsmath}		
\usepackage{amssymb}		
\usepackage{autonum}
\usepackage{listings}
\usepackage[linesnumbered,ruled,vlined]{algorithm2e}
\usepackage{comment}
\usepackage{subfigure}
\usepackage{graphicx}
\usepackage{balance}
\usepackage{flushend}

\DeclareMathOperator*{\argmax}{arg\,max}

\begin{document}

\date{}

\title{\Large \bf Model-Centric and Data-Centric Aspects of Active Learning \\  for Deep Neural Networks}

\author{
{\rm John Daniel Boss\'er}\\
\normalsize Chalmers University of Technology, Link\"oping University, \texttt{daniel.bosser@liu.se}
\and
{\rm Erik S\"orstadius}\\
\normalsize Chalmers University of Technology, \texttt{eriksoerstadius@gmail.com}
\and
{\rm Morteza Haghir Chehreghani}\\
\normalsize Chalmers University of Technology, \texttt{morteza.chehreghani@chalmers.se}
\vspace{5mm}
}

\maketitle

\begin{abstract}
We study different aspects of active learning with deep neural networks in a consistent and unified way.  i) We investigate incremental and cumulative training modes which specify how the newly labeled data are used for training. ii) We study  active learning w.r.t. the model configurations such as the number of epochs and neurons as well as the choice of batch size. iii) We consider in detail the behavior of query strategies and their corresponding informativeness measures and accordingly propose more efficient querying procedures. iv) We perform statistical analyses, e.g., on actively learned classes and test error estimation, that reveal several insights about active learning. v) We investigate how active learning with neural networks can benefit from pseudo-labels as proxies for actual labels.
\end{abstract}

\section{Introduction}

The recent advances in Artificial Intelligence (AI) have been focused on Artificial Neural Network (ANN) models \cite{LeCunBH15,Goodfellow-et-al-2016} with great success in pattern recognition (computer vision, NLP) and some reinforcement learning applications (e.g., AlphaGo \cite{SilverHMGSDSAPL16}). These models have become popular due to their high flexibility, capacity and accuracy. However, they are data-hungry, i.e., they require a huge amount of labeled data for training. 
In an image classification task, for instance, the model needs a large training data set possibly in the order of millions images \cite{Goodfellow-et-al-2016}. In practice, many applications and tasks might not have access to such a large training data set. In particular, annotating and labeling data is often time consuming, tedious and expensive. 
Usually, we cannot afford to annotate all data, due to a limited budget, and the goal is thus to attain the best gain from this limitation.

The paradigm of choosing the most informative data to label is referred to as \emph{active learning} \cite{CohnGJ96,settles.tr09}.
The main challenge is that there is no objectively superior way to determine how informative each data label is.  Therefore, active learning is usually performed in the context of sequential decision making, where at every step, two operations are performed: i) choose the next object or data to be labeled and annotated, ii) update the underlying (training) model w.r.t. the new annotation. The next object for labeling is chosen according to a  \emph{query strategy} or an \emph{acquisition function}. The objective is to  maximize the test accuracy or minimize its loss, with  a minimal number of queries. Common query strategies aim to reduce uncertainty w.r.t. objectives such as \textit{margin}, \textit{variance}, \textit{uncertainty}, \textit{model change} and \textit{entropy} \cite{settles.tr09,WilsonMaximizingOptimization,Sener2017ActiveApproach}.
The method in \cite{Houlsby2011BayesianLearning} applies predictive entropy to the Gaussian Process
Classifier to develop a Bayesian approach for active learning called BALD. This method has been investigated in the context of deep learning too, using some approximate techniques based on drop-out \cite{GalIG17,BatchBALD19}.

The queries may be performed in the form of class labels or pairwise relations. Querying class labels is usually more common \cite{CohnGJ96,DasguptaHM07,GalIG17, Hanneke07}, which has been extensively used in robotics, text and image classification, drug discovery, troubleshooting and log data analysis \cite{tong2001active,Jarl2021,YanCJ18,ChenRCK17}.
Querying the pairwise relations has been mainly studied in the context of semi-supervised learning and (supervised/interactive) clustering \cite{NIPS2016_6449,AwasthiZ10}. Some of such methods essentially adapt the clustering models on positive and negative edge weights such as correlation clustering and shifted min cut \cite{BansalBC04,ChehreghaniICDM17}.
Noisy active learning has been developed separately for both label-based annotations \cite{YanCJ16,NaghshvarJC13,FrenayIEEE2014} and pairwise annotations \cite{NIPS2017_Mazumdar}, with applications in different areas such as recommendation systems, computer vision, and medicine.

In this paper, we investigate several novel data-centric and model-centric aspects of active learning with deep neural network models. i) We investigate incremental and cumulative training modes which specify how the available labeled data set is used to train the neural network model.
ii) We investigate in detail the behaviour of query strategies (acquisition functions) and their relations to training and test performance.
iii) Neural networks are models with a large capacity. Thus, we study how active learning depends on different configurations such as the number of epochs and neurons as well as the choice of batch size.
iv) We consider the cases where  the acquisition function can be investigated only on a subset (sample) of the entire unlabeled data. This can be due to computational limitations, or when  not all the unlabeled data is available right from at the beginning. We study the behaviour of query strategies in this setting.
v) We study the initial bias a query strategy like entropy might induce when the initial network is not trained properly. We suggest random start to address this issue.
vi) We develop a semi-supervised active learning paradigm by extending the method in \cite{pseudo-label-lee-hyun} which enables the model to employ the unlabeled data for a better training, in addition to the labeled data.
These studies help us for a better understanding of active learning for neural network models and provide useful guidelines for more effective design of such paradigms.

\section{Background}

We are given a set of objects with indices $\{i\}$ and the respective representations (features) $\{\mathbf{x}_i\}$. The true label of object $i$ is indicated by $y_i$ which might be given or not.  Thus, we consider two data sets $L$ and $U$, where $L$ includes the object whose true labels are known whereas $U$ consists of unlabeled objects. We may consider a separate test data set to evaluate and compute the test accuracy of the trained models.
\newcommand{\x}{\mathbf{x}}
Given an input $\mathbf{x}_i$ to a neural network $C$, we obtain the softmax outputs (predictive class probabilities) $\hat{y}_{ic}$ for object $i$. $\hat{y}_{ic}$ indicates the  probabilistic prediction $P_C(\hat{y} = c | \x_i)$, i.e., the probability that $i$ belongs to class $c$. A query strategy $A$ assigns an informativeness measure $I^A_i$ to every $i$ in the unlabeled data set $U$. The query strategy then sorts the unlabeled objects based on $I_i^A$ and selects a batch $B$ consisting of $n$ objects with the highest $I^A_i$ ($n$ is the batch size and is fixed in advance).
\newcommand{\nb}{n}
\begin{align}
    B = \{B_1,...,B_{\nb}\}, \texttt{ s.t. } & I^A_{B_{1}} \ge I^A_{B_{2}} \ge \dots \ge I^A_{B_{\nb}} \\  \texttt{ and } & I^A_{B_{\nb}} \ge I^A_i \;\; \forall i \in U \setminus B.
\end{align}
For each object in $B$, we ask the oracle (that can be for example an expert) to provide a class label. We then update the set of all labeled data by $L \leftarrow L\cup B$ and the set of all unlabeled data by $U \leftarrow U \setminus B$.
\newcommand{\MIM}{\bar{I}}Sometimes we might need to compute the mean informativeness measure $\MIM_B^A$ of the query strategy $A$ for the selected batch $B$ as
   $\MIM_B^A = \frac{1}{|B|} \sum_{i \in B} I^A_{i}.$

The main query strategies used this paper, i.e. \textit{Random}, \textit{Margin}, and \textit{Entropy}, only differ in the way they assign their corresponding informativeness measures.

\textbf{Random (R)}
\newcommand{\unif}{\mathrm{unif}}
\label{sec:theory:query_strategies:random}
assigns a uniformly distributed informativeness measure $I^R_{i} \sim \unif(0, 1)$ to all $i \in U$.

\textbf{Margin (M)}
\label{sec:theory:query_strategies:margin}
computes informativeness measures for each object $i \in U$ as $I^M_{i} = -[P_C(\hat{y} = \tilde{c}_1 | \mathbf{x}_i) - P_C(\hat{y} = \tilde{c}_2 | \mathbf{x}_i)]$,
where $\tilde{c}_1$ and $\tilde{c}_2$ are the most probable and the second most probable classes predicted by classifier $C$ for $i \in U$.

\textbf{Entropy (E)}
assigns the informativeness measure based on the entropy of the predictive distribution, i.e., $I^E_{i} = -\sum_{c} P_C(\hat{y} = c | \mathbf{x}_i) \log P_C(\hat{y} = c | \mathbf{x}_i)$.

\paragraph*{Neural network architectures and models}
We use the MNIST \cite{lecun-mnisthandwrittendigit-2010}, Fashion-MNIST \cite{DBLP:journals/corr/abs-1708-07747} and CIFAR-10 \cite{Krizhevsky2009LearningImages} data sets for training and evaluation of the networks in different active learning paradigms. There are 70000 images in MNIST and Fashion-MNIST, where each image consists of $28\times 28$ features. CIFAR-10 contains 60000 images with $32 \times 32 \times 3$ features. 10000 images from each data set are reserved as a test set. 
The same network is used for both the MNIST and Fashion-MNIST data sets and is referred to as $\mathrm{ANN1}_{m,e}$. The network consists of three layers with $m$ neurons in the hidden layer and $e$ is the number of epochs.
For the CIFAR-10 data set, we use a convolutional neural network consisting of several layers that uses 50 epochs, which we refer to as $\mathrm{CNN1}$.

\section{Query Strategies and Training Modes}
\label{sec:background:query_strategies_and_training_methods}

\paragraph*{Methodology} Design of an effective query strategy constitutes a core component of any active learning paradigm.
Different query strategies have been well investigated in several studies, for example in \cite{settles.tr09,WilsonMaximizingOptimization,Sener2017ActiveApproach}.
However, after querying the label of a new object, the underlying model $C$ should be updated accordingly. In this section, we study two modes of updating the model, where we call them \emph{incremental training} and \emph{cumulative training}.

\begin{itemize}
\item \emph{Incremental training}: We consider the trained model on the previously labeled objects $L$ and update its parameters (i.e., the weight and bias parameters of the neural network) only using the new labeled batch $B$.

\item  \emph{Cumulative training}: We re-initialize the weight and bias parameters of the model and use $L\cup B$ for training.
\end{itemize}

Incremental training is usually faster and computationally more efficient. 
However, the incremental mode induces a certain order on the way the labeled data is fed into the network which might yield some bias.
The cumulative method, on the other hand, utilizes the entire labeled data set at each acquisition step and thus induces randomness to the training order. Computational runtime might not be the main objective in active learning as its goal is to minimize the number of queries to the oracle.

Active learning methods usually use the incremental training mode. The cumulative training has not been well studied in this context. In this section, we study combinations of query strategies and training modes for a large neural network model. We use $\mathrm{ANN1}_{100,1}$ neural networks to perform the experiments on MNIST and Fashion-MNIST data sets. We choose a batch size of 120 objects. 
We  also investigate the effects of selecting data with one network, and training another network with the already selected data. These two networks, i.e., the \textit{selection network} and \textit{evaluation network}, might differ in the number of epochs and neurons. This separation of networks helps us to study if the difference in results is due to the specific training modes, or it is because of the different labeled data selected in each of the settings.

\begin{figure*}[thb]
    \centering
    \hspace{-5mm}
    \subfigure[MNIST.]
    {
        \includegraphics[width=0.25\textwidth]{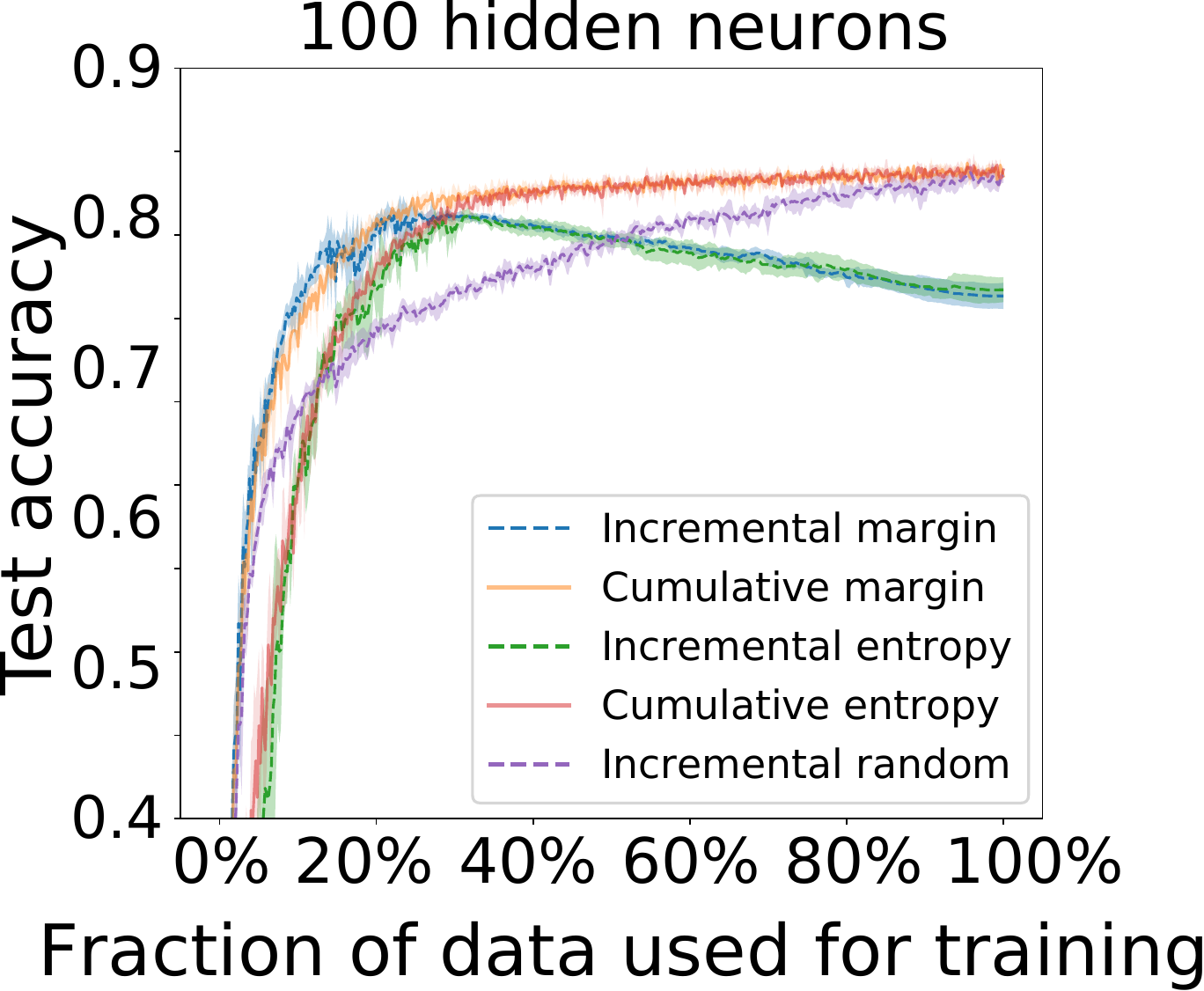}
        \label{fig:query_strategies:medium:mnist}
    }
    \hspace{-3mm}
    \subfigure[Fashion-MNIST.]
    {
        \includegraphics[width=0.25\textwidth]{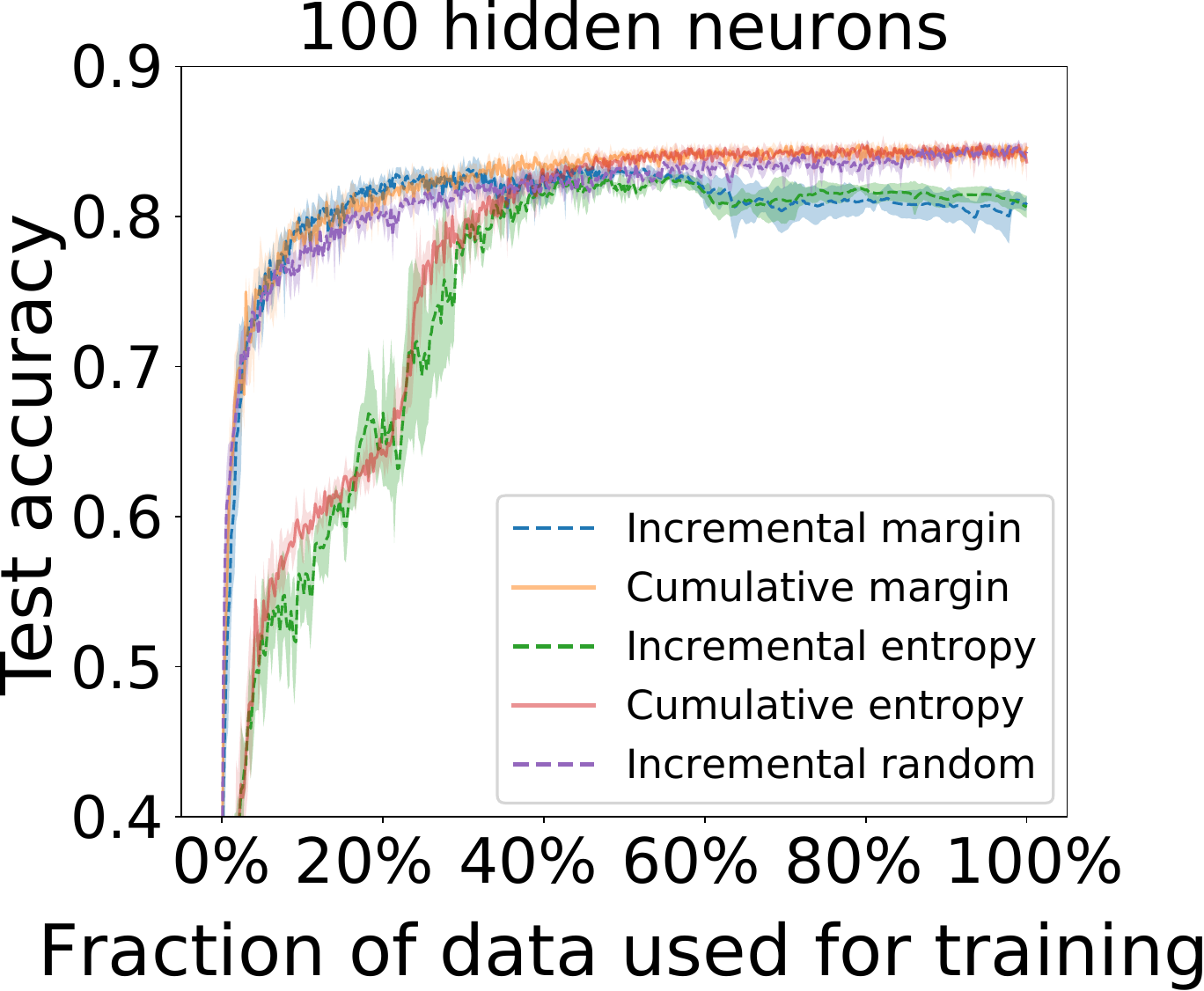}
        \label{fig:query_strategies:medium:fashion}
    }
    \hspace{-5mm}
    \subfigure[Varying epochs.]
    {
        \includegraphics[width=0.25\textwidth]{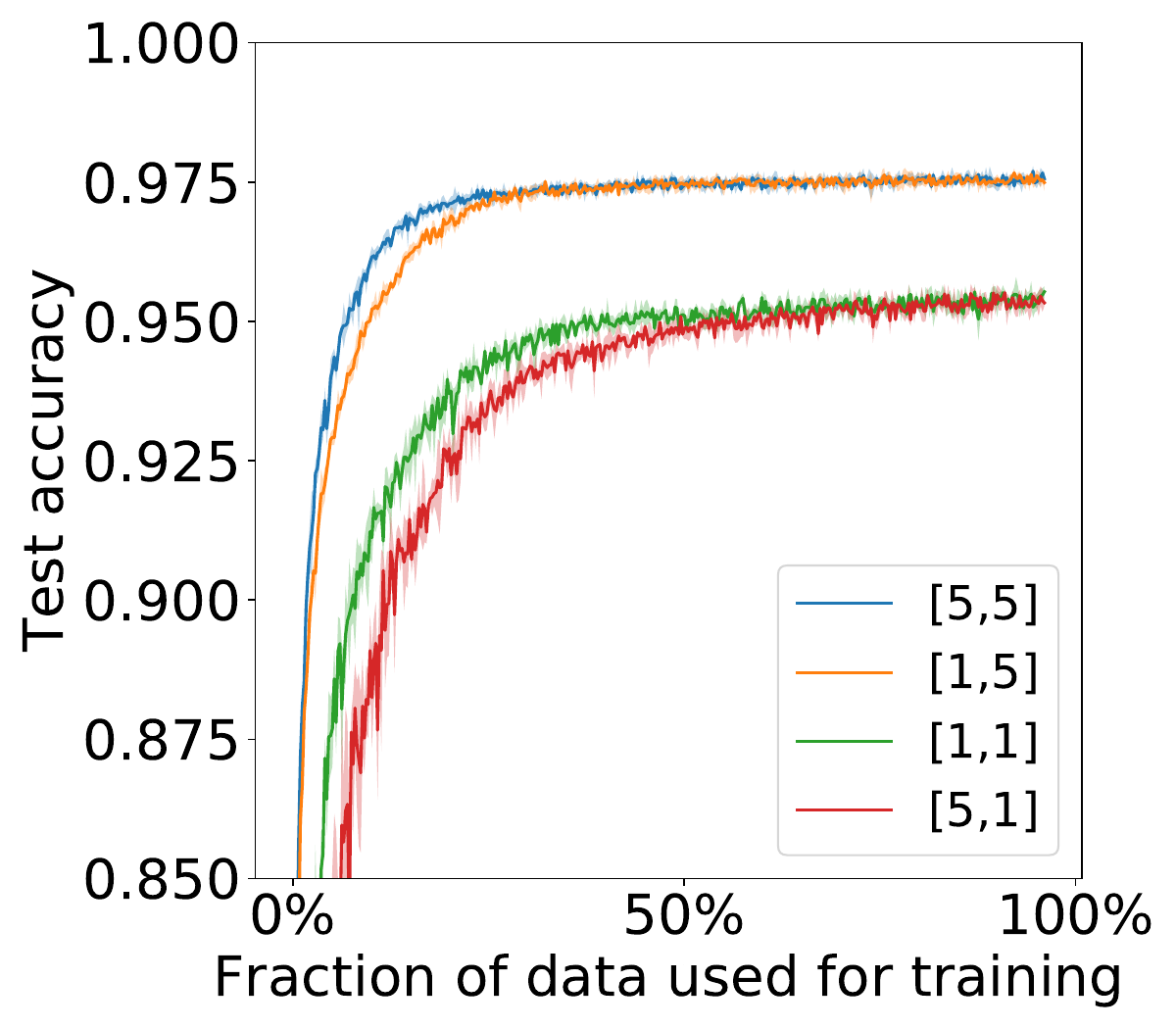}
        \label{fig:evaluation_of_sampled_data_sets:varying_epochs_cumulative_epochs_x_x:mnist}
    }
    \hspace{-5mm}
    \subfigure[Varying hidden neurons.]
    {
        \includegraphics[width=0.25\textwidth]{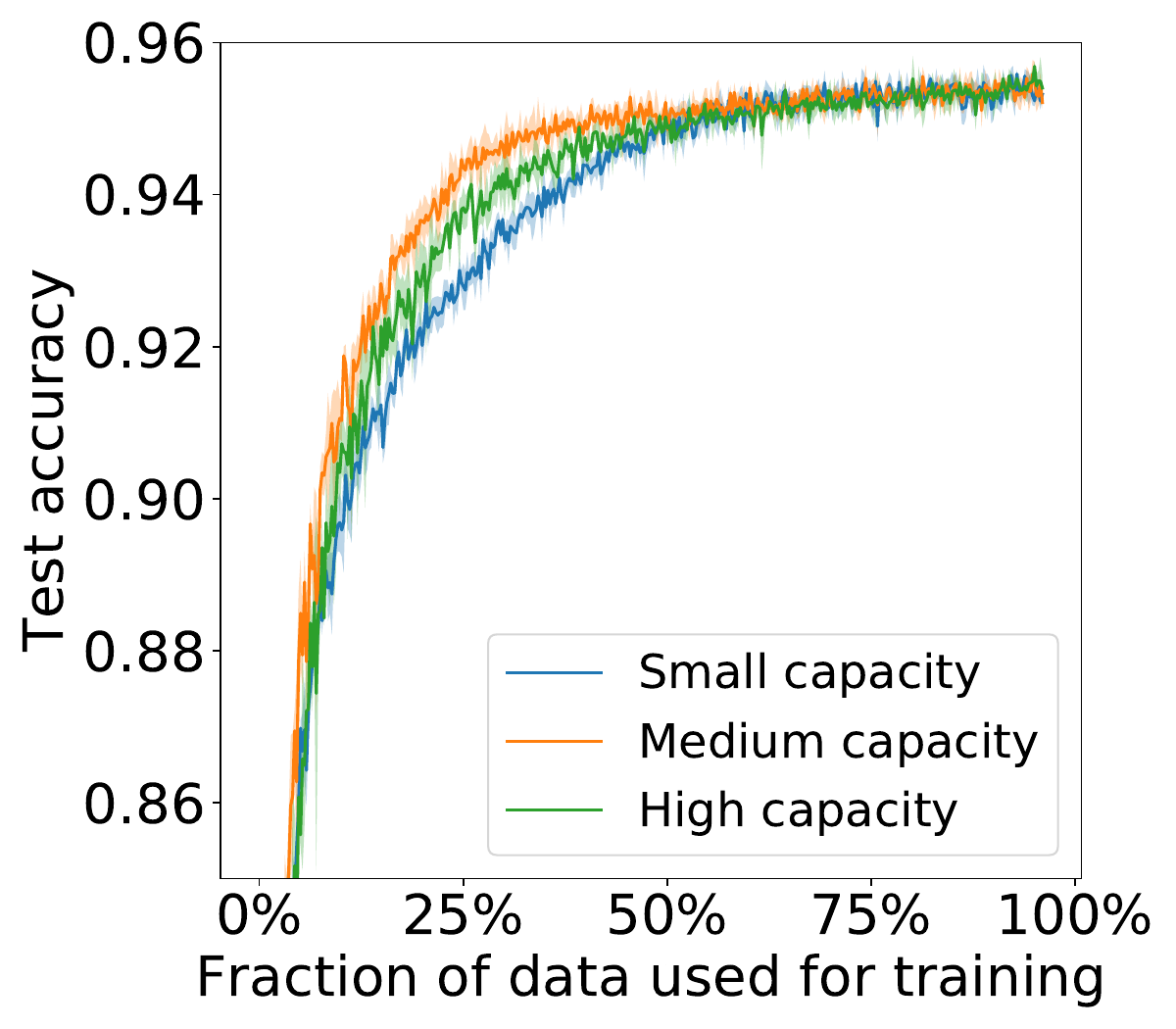}
        \label{fig:evaluation_of_sampled_data_sets:cumulative_varying_number_of_neurons:mnist}
    }
    \caption{Test accuracy for $\mathrm{ANN1}_{100,1}$ network on MNIST (a) and Fashion-MNIST (b). In (c) we vary the number of epochs in the selection and evaluation network. Figure (d) shows the test accuracy for varying number of hidden neurons in the selection network. The results in (c) and (d) are on MNIST.
    }
    \label{fig:query_strategies:medium:mnist:fashion}
\end{figure*}

\paragraph*{Results and discussion}
\label{sec:results_and_discussion:query_strategies_and_training_methods}
Figures \ref{fig:query_strategies:medium:mnist} and \ref{fig:query_strategies:medium:fashion} show the test accuracy of the networks for different ratios of the labeled data, where 100\% corresponds to the case where the whole unlabeled data set of 60000 images has been labeled and used for training. We report the results for different choices of query strategies and training modes.
With random query strategy, the results for the cumulative and incremental training modes are very similar, thus we report only the results of incremental mode.

In Figures \ref{fig:query_strategies:medium:mnist} and \ref{fig:query_strategies:medium:fashion} we observe that the margin query strategy initially outperforms both the random and the entropy query strategies in both of the training modes.
This observation is consistent with the prior studies with other models than neural network \cite{margin_performs_well}. Comparison of query studies has been studied in several other works with consistent observations, thus we will not focus a lot in this paper. Thereby, in most of our studies, we will use the margin query strategy.
In addition, we observe that the entropy query strategy performs similarly and sometimes even worse than the random strategy at the beginning, which might be due to the bias induced by this specific query strategy. In Section \ref{sec:background:random_start} we study this effect and a method to mitigate it.

An even more important observation is that the cumulative training performs better than the incremental training, in particular when the size of the labeled data is not too small. For the MNIST data set, the incremental and the cumulative modes perform similarly up to 10000 images labeled and used in training. After that, cumulative mode starts outperforming. The networks trained in the incremental mode experience a decline in test accuracy after a certain number of objects have been labeled.  This is probably due to the bias induced by the order of the data processing that the incremental mode yields.
The training order is randomized with the cumulative mode which has access to all the labeled data randomly via stochastic gradient descent.

In summary, margin query strategy with incremental training outperforms the alternatives initially. The cumulative mode yields better results throughout the training process, in particular when a large fraction of the data has been already labeled.

Figure \ref{fig:evaluation_of_sampled_data_sets:varying_epochs_cumulative_epochs_x_x:mnist} shows the test accuracy of the evaluation network while varying the number of epochs in the networks. The legend $[a,b]$ means that $\mathrm{ANN1}_{100, a}$ and $\mathrm{ANN1}_{100, b}$ are used as the selection and evaluation networks respectively. Figure \ref{fig:evaluation_of_sampled_data_sets:cumulative_varying_number_of_neurons:mnist} shows the test accuracy of the $\mathrm{ANN1}_{100,1}$ evaluation network. The selection network is either $\mathrm{ANN1}_{10,1}$ (Small), $\mathrm{ANN1}_{100,1}$ (Medium) or $\mathrm{ANN1}_{1000,1}$ (High). Both Figures \ref{fig:evaluation_of_sampled_data_sets:cumulative_varying_number_of_neurons:mnist} and \ref{fig:evaluation_of_sampled_data_sets:varying_epochs_cumulative_epochs_x_x:mnist} imply that it is always beneficial to have the same number of epochs and neurons for the selection and evaluation networks. Therefore, the evaluation network should be decided before selecting data, since the selection and evaluation networks should have the same number of epochs and neurons.

\begin{figure}
    \centering
    \subfigure[MNIST.]{\includegraphics[width=0.22\textwidth]{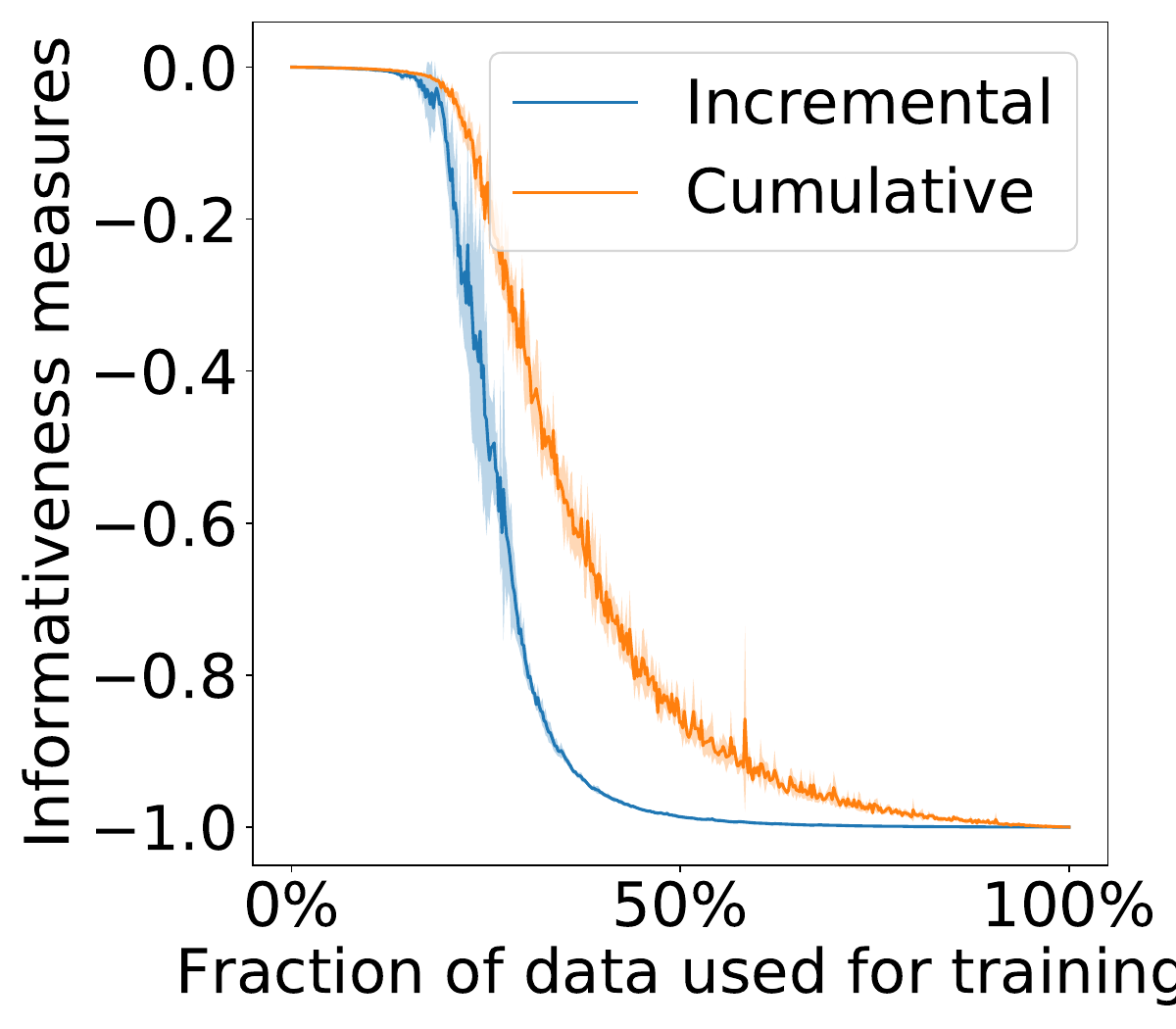}}
    \subfigure[Fashion-MNIST.]{\includegraphics[width=0.22\textwidth]{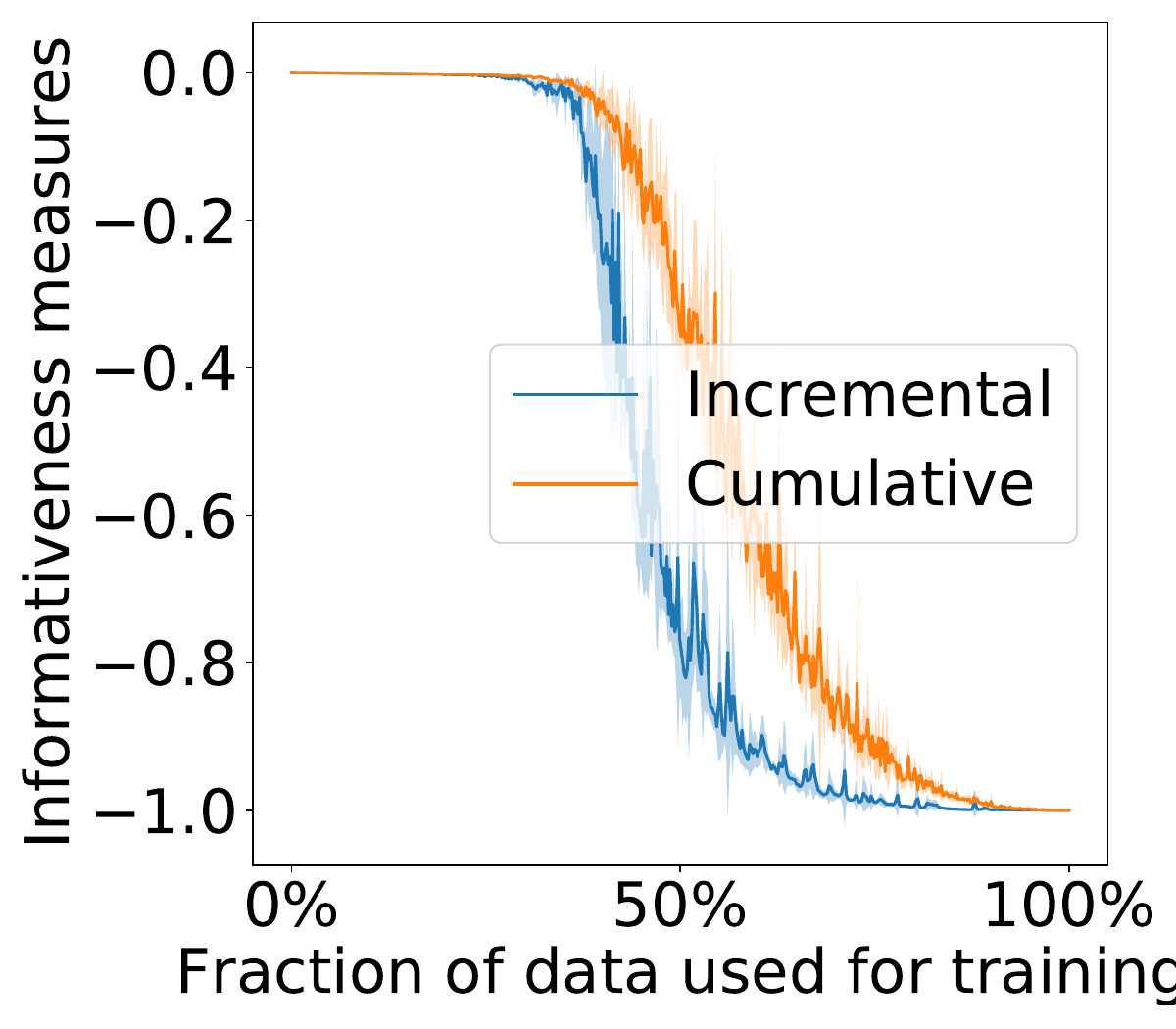}}
    \caption{How the mean informativeness measure $\MIM^M_B$ changes depending on training mode. }
    \label{fig:query_strategies_and_training_methods:inc_vs_cum:infor_measure}
\end{figure}

Figure \ref{fig:query_strategies_and_training_methods:inc_vs_cum:infor_measure} illustrates the $\MIM^M_B$ of the selected batches for both incremental and cumulative training modes. These results correspond to the results shown in Figures \ref{fig:query_strategies:medium:mnist} and \ref{fig:query_strategies:medium:fashion}. We observe that $\MIM^M_B$ decreases after 20\% of all MNIST data has been used in training, or at 40\% in the Fashion-MNIST case. Moreover, $\MIM^M_B$ approaches -1 earlier when incremental mode is used. It is also approximately at the mentioned percentages where we see a decrease in test accuracy in Figure \ref{fig:query_strategies:medium:mnist} and \ref{fig:query_strategies:medium:fashion} in the incremental margin case.
Thus, a network trained in incremental mode trains on more batches with low $\MIM_B^M$. This observation opens up for the hypothesis that the incremental training yields training with non-informative data which in turn causes the decrease of the test accuracy.

\begin{figure}[htb!]
    \subfigure[MNIST.]{
        \includegraphics[width = 0.22\textwidth]{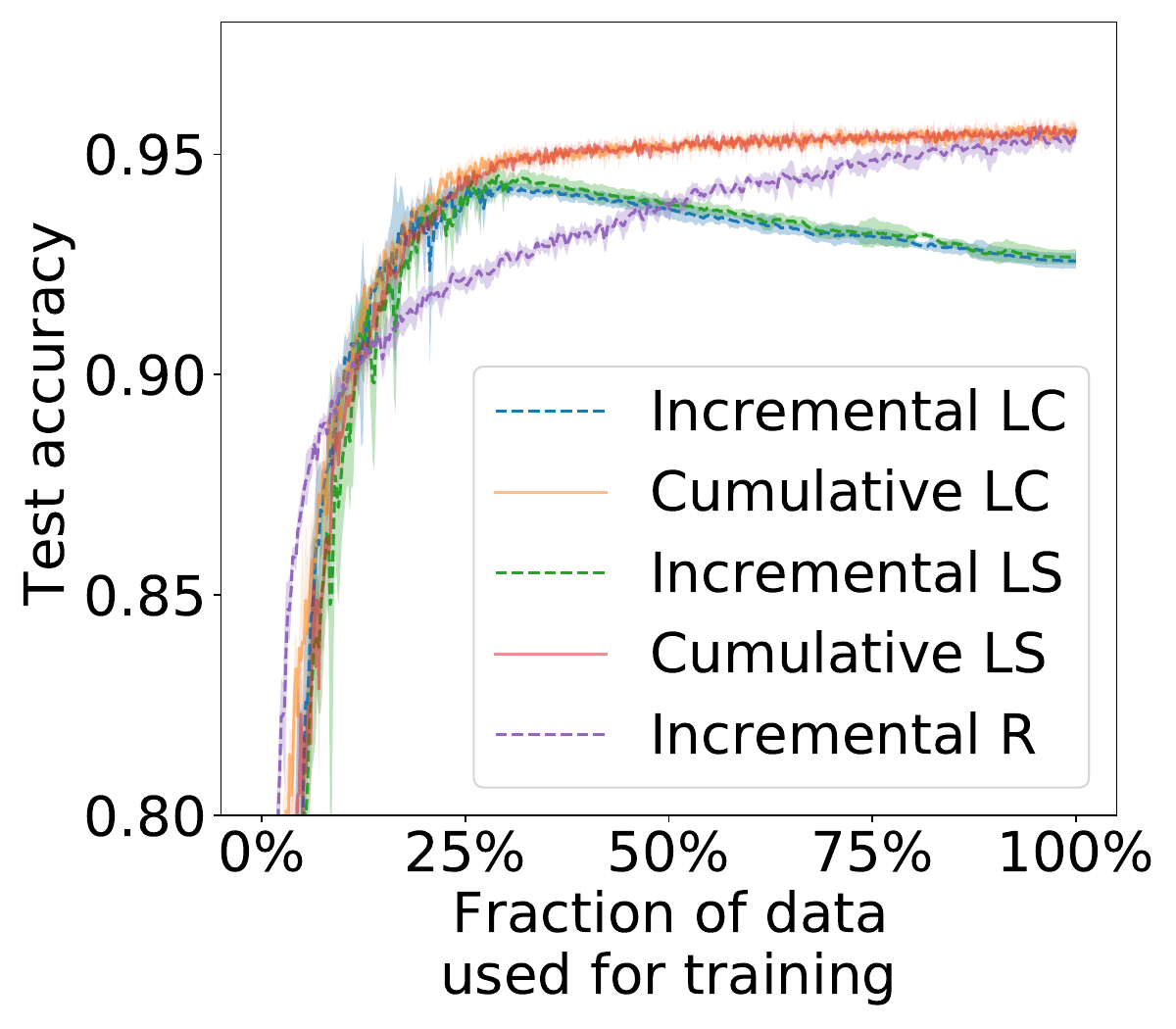}
    }
    \subfigure[Fashion-MNIST.]{
        \includegraphics[width = 0.22\textwidth]{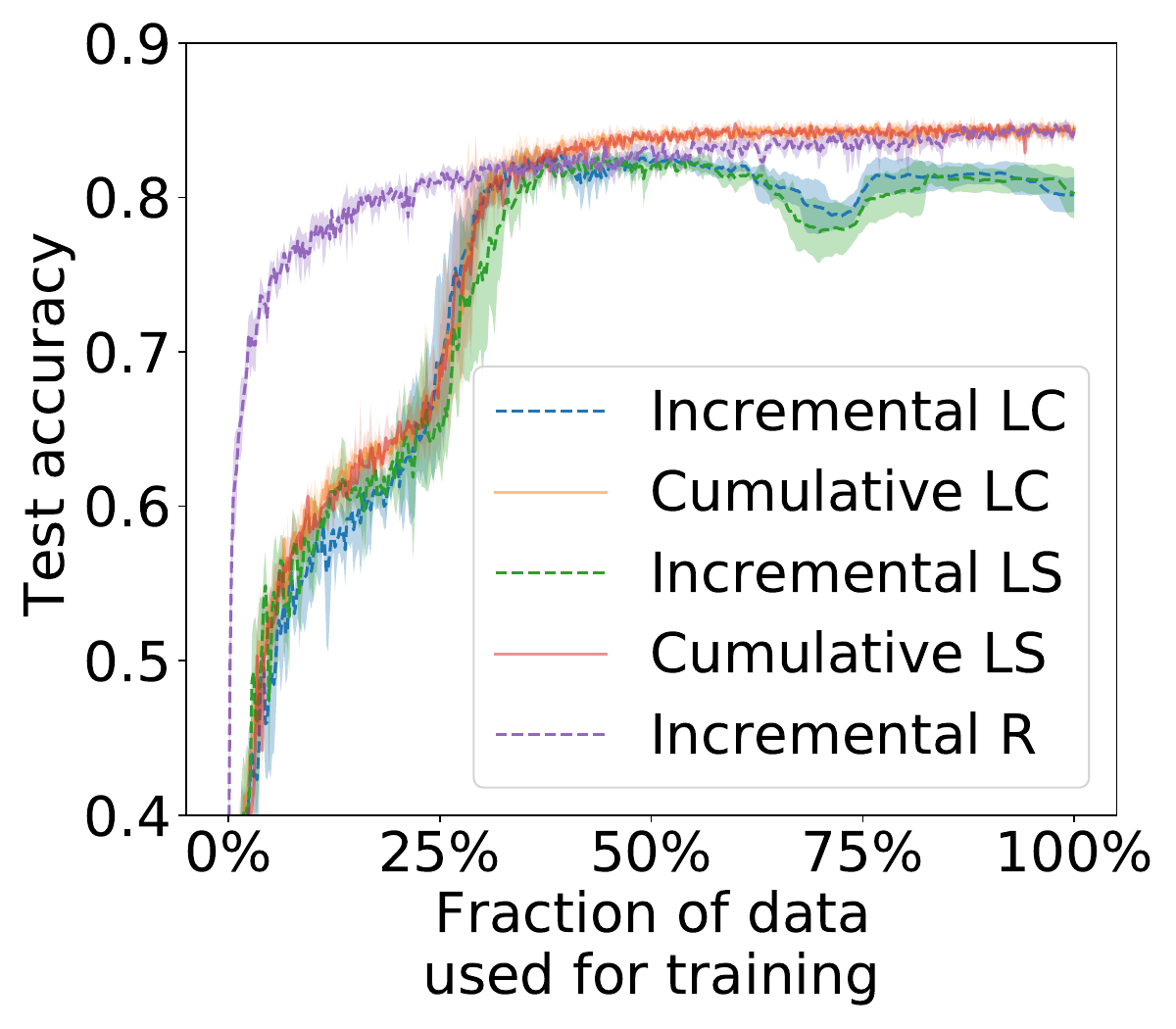}
    }
    \caption{Test accuracy of $\mathrm{ANN1}_{100, 1}$ on the MNIST (a) and Fashion-MNIST (b) data sets, with the least confident and least squares query strategies. }
    \label{fig:app:additional_qs}
\end{figure}

Finally, as a side study and in addition to the query strategies in Section \ref{sec:background:query_strategies_and_training_methods}, we examine the performance of two additional query strategies: the \textit{Least Confident} (LC), and \textit{Least Squares} (LS) strategies, with the corresponding informativeness measures $I^{LC}_i = -\max_c P_C(y = c | \mathbf{x}_i)$ and $I^{LS}_i = - \sum_{c} [ P_C(y = c | \mathbf{x}_i) - \frac{1}{N_c} ]^2$. Figure \ref{fig:app:additional_qs} shows their performance in relation to the random query strategy using the cumulative and incremental training modes.
We observe that the performance of $LC$ and $LS$ is similar to the performance of $E$ in Figure \ref{fig:query_strategies:medium:mnist:fashion}.

\section{Network Capacity}

\label{sec:background:capacity_of_the_network}
\paragraph*{Methodology} In Section \ref{sec:background:query_strategies_and_training_methods}, we saw that the margin query strategy yields the best performance for the $\mathrm{ANN1}_{100,1}$ network. Here, we investigate the different query strategies and training modes on networks of different capacities.
The methodology of performing the experiments is consistent with the study in Section \ref{sec:background:query_strategies_and_training_methods}.

\paragraph*{Results and discussion}

\begin{figure}[htb!]
    \centering
    \subfigure[10 hidden neurons, MNIST.]{
        \includegraphics[width = 0.22\textwidth]{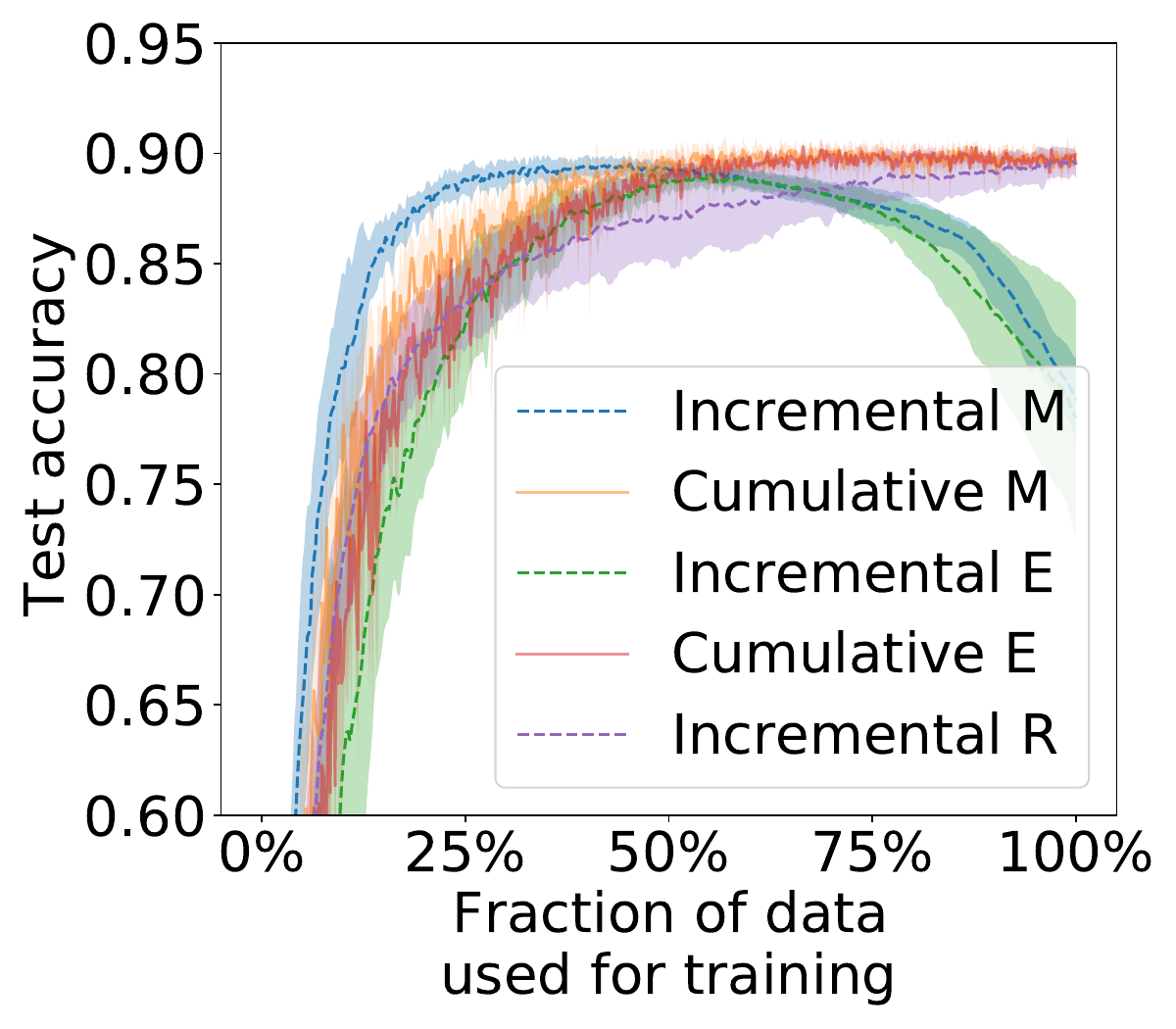}
    }
    \subfigure[1000 hidden neurons, MNIST.]{
        \includegraphics[width = 0.22\textwidth]{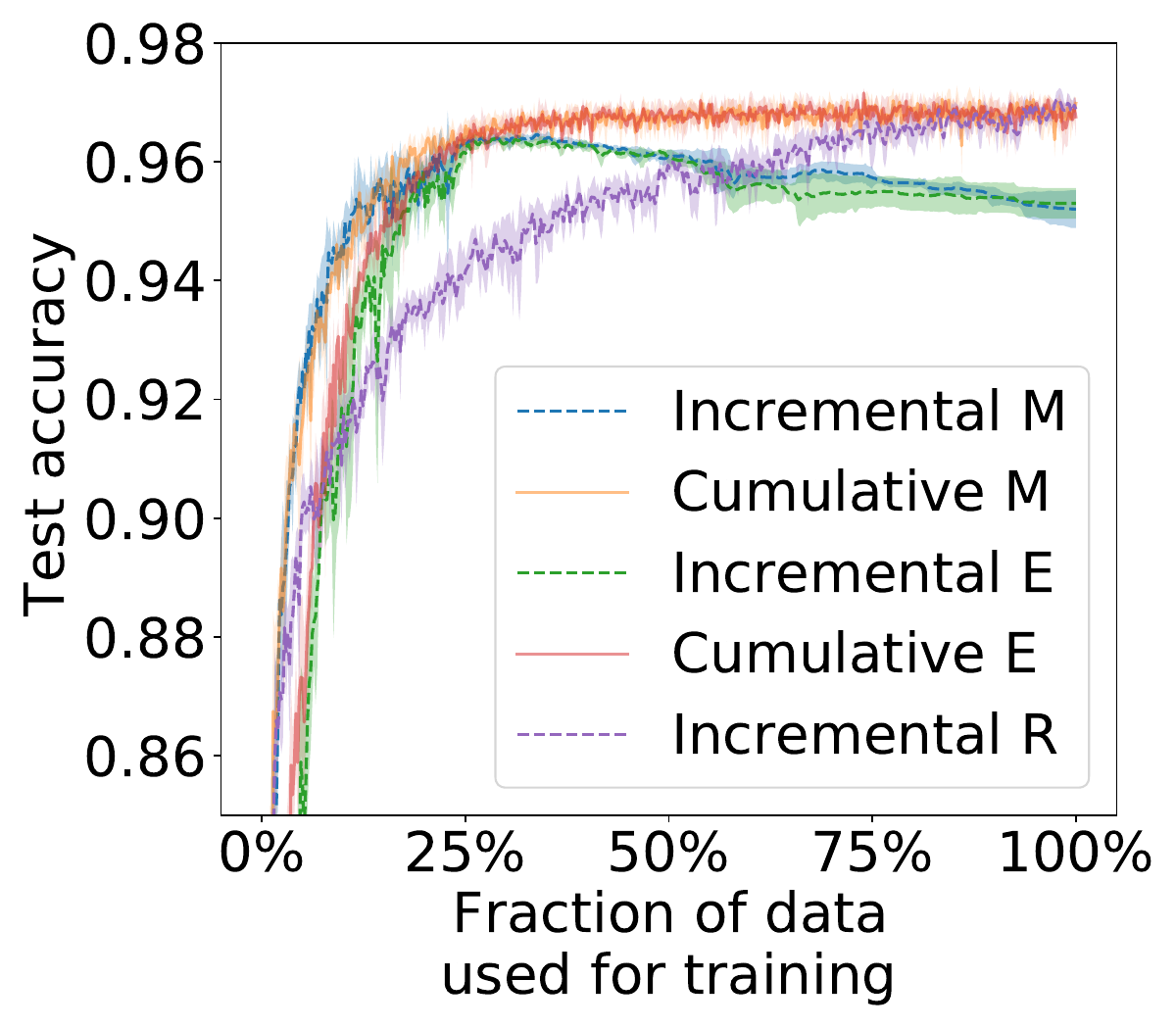}
    }

    \subfigure[10 hidden neurons, Fashion-MNIST.]{
        \includegraphics[width = 0.22\textwidth]{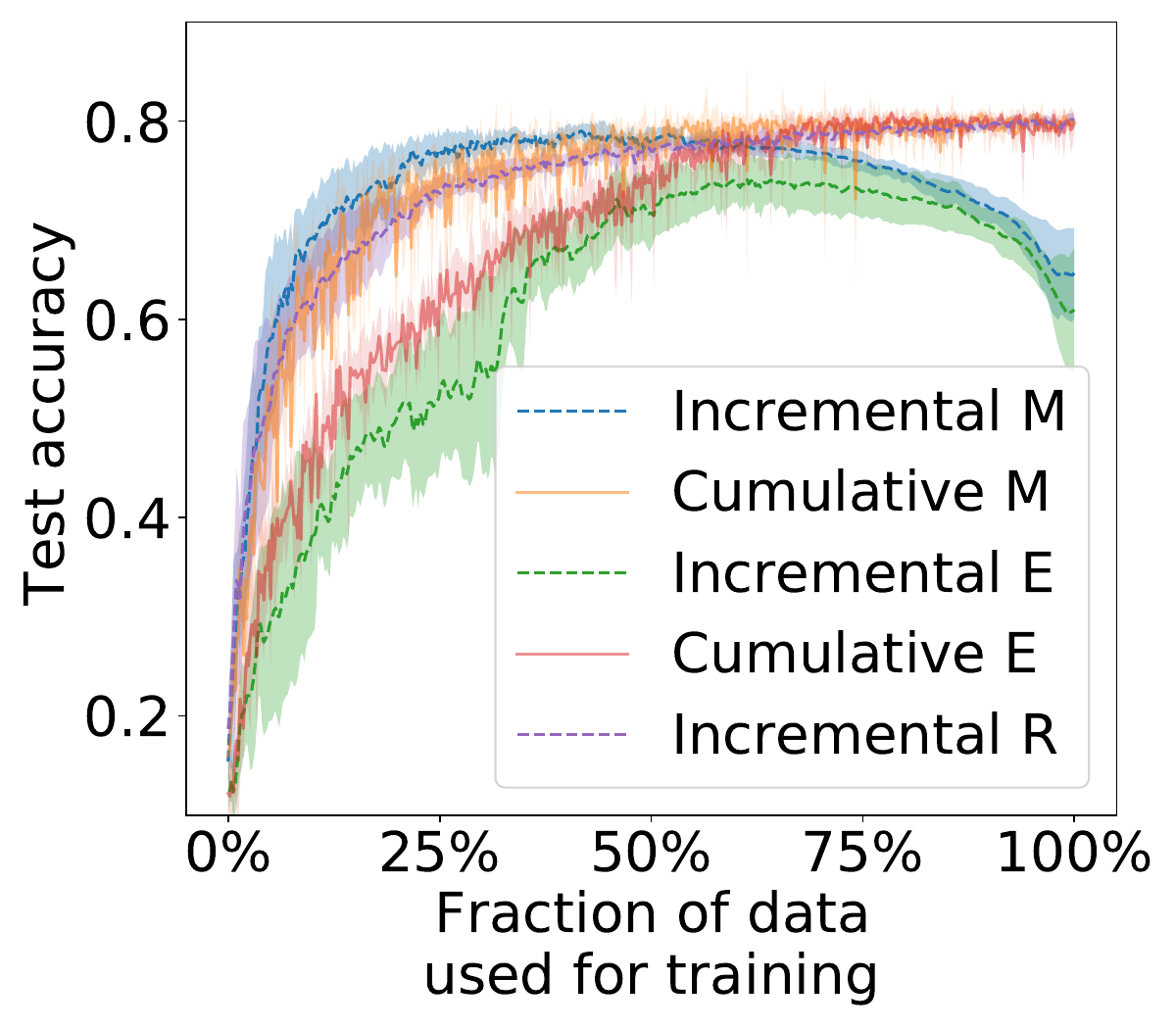}
    }
    \subfigure[1000 hidden neurons, Fashion-MNIST.]{
        \includegraphics[width = 0.22\textwidth]{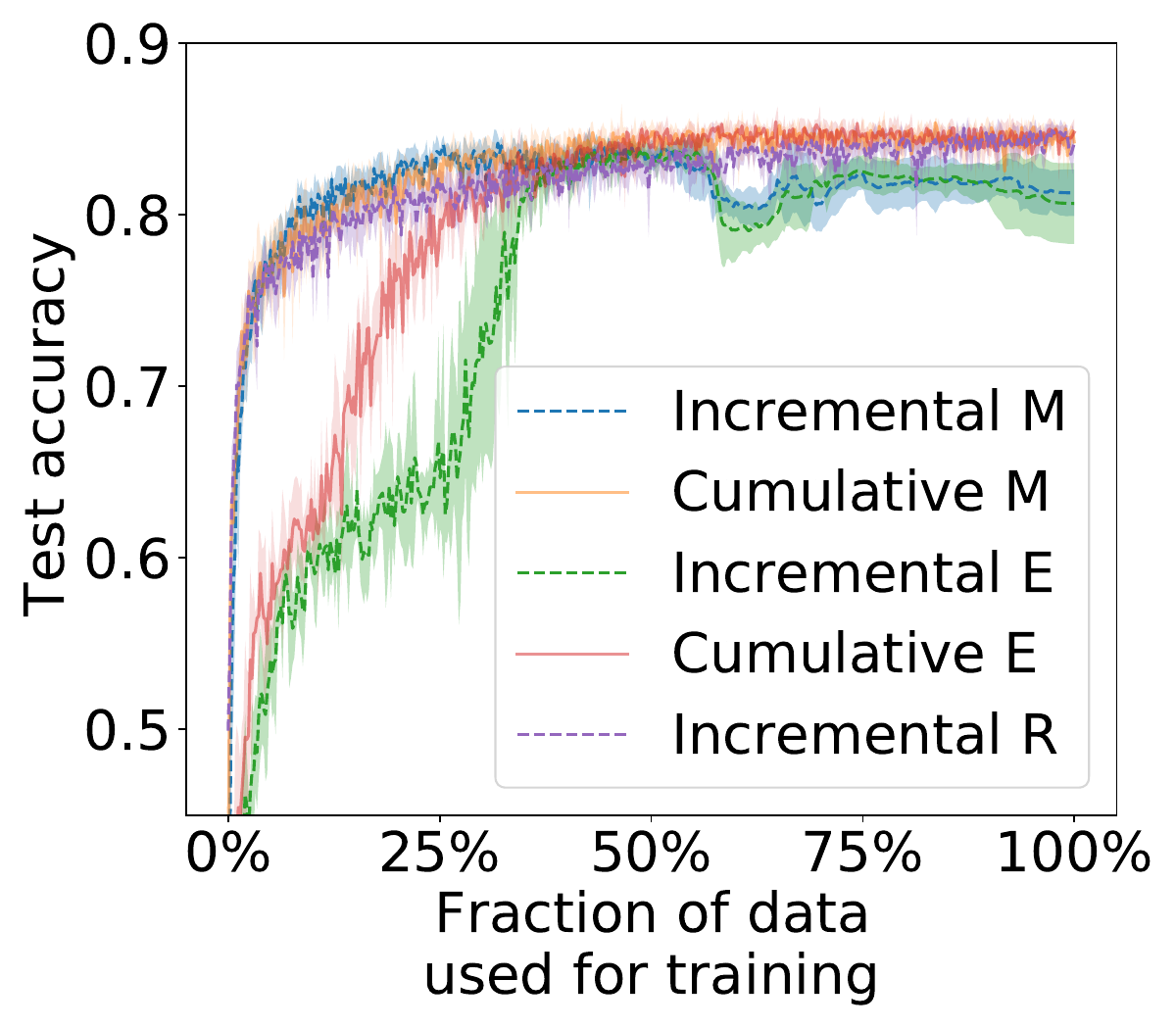}
    }
    \caption{Performance of the random, margin, and entropy query strategies on $\mathrm{ANN1}_{10,1}$ and $\mathrm{ANN1}_{1000,1}$ networks with incremental and cumulative training modes. }
    \label{fig:results:capacities}
\end{figure}

Figure \ref{fig:results:capacities} shows the test accuracy of $\mathrm{ANN1}_{10,1}$ and $\mathrm{ANN1}_{1000,1}$ networks trained with the margin and entropy query strategies, where the batch size is 120. We report the results for both the incremental and cumulative training modes both on the MNIST and Fashion-MNIST data sets.
The results on $\mathrm{ANN1}_{100,1}$ are those reported in Figure \ref{fig:query_strategies:medium:mnist:fashion}.
In general, the results are consistent for different networks of varying capacities. The margin query strategy, trained with  cumulative training, performs as good as or better than the random query strategy. It consistently outperforms the entropy query strategy.

The comparison between the incremental and cumulative training modes demonstrates consistent results: i) we  observe a decline of the test accuracy of the networks trained with the incremental training mode, while the test accuracy of the networks trained with the cumulative mode increases as more data is fed to the network.  ii) However, the incremental training mode may have an advantage at the early steps of the training which yields comparable or even better (with $\mathrm{ANN1}_{10,1}$) results compared to cumulative training. This might explain why incremental training is used more often in practice, since active learning is sometimes concerned with labeling a small fraction of the total data.

When the capacity of the network is small, i.e., on $\mathrm{ANN1}_{10,1}$, the difference between the random and the margin query strategies is small for the cumulative training.
Moreover, the difference in test accuracy between the incremental and cumulative training is more prominent for a smaller network, i.e.,  the decline of the test accuracy for a smaller network is more severe when using incremental training. The reason could be that a network with a smaller capacity is more sensitive to the bias induced by the training  order caused by the specific query strategy. A larger network is more flexible due to its larger capacity.

\section{Choice of Batch Size}

\paragraph*{Methodology}
The \textit{batch size} $n_B = |B|$ is the number of object labels queried together before (re)training the network at the next step. Ideally, to ensure that the most informative samples are labeled, only  one object label should be queried per training step, i.e. $n_B = 1$. However, this slows down the training process. Instead, $n_B$ samples with the highest informative measures are selected. 
We thus study how $n_B > 1$ affects the performance of the trained network for both the incremental and cumulative training modes.

\paragraph*{Results and Discussion}

\begin{figure}
\centering
\subfigure[Cumulative MNIST.]{
    \includegraphics[width=0.22\textwidth]{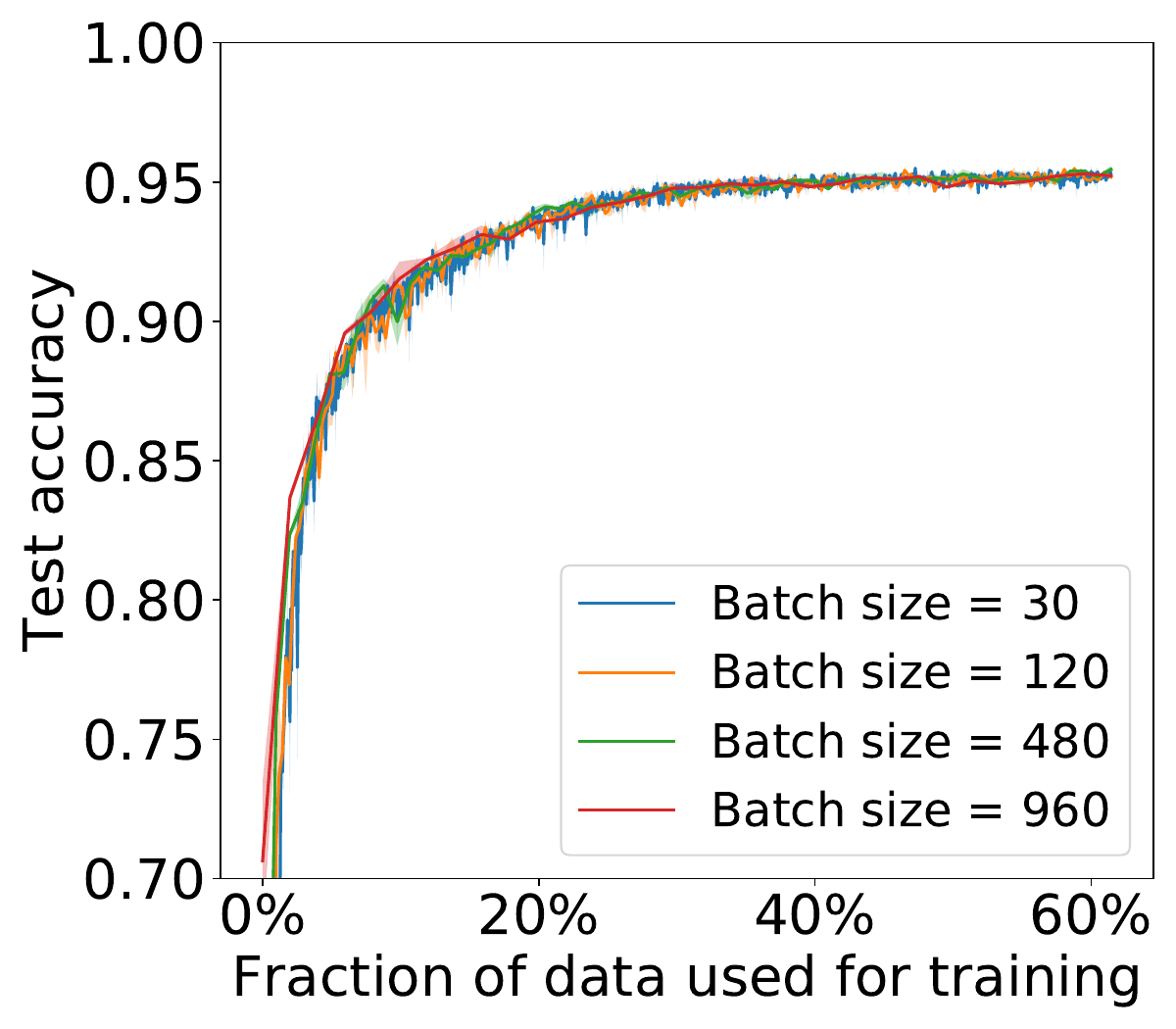}
}
\subfigure[Incremental MNIST.]{
    \includegraphics[width=0.22\textwidth]{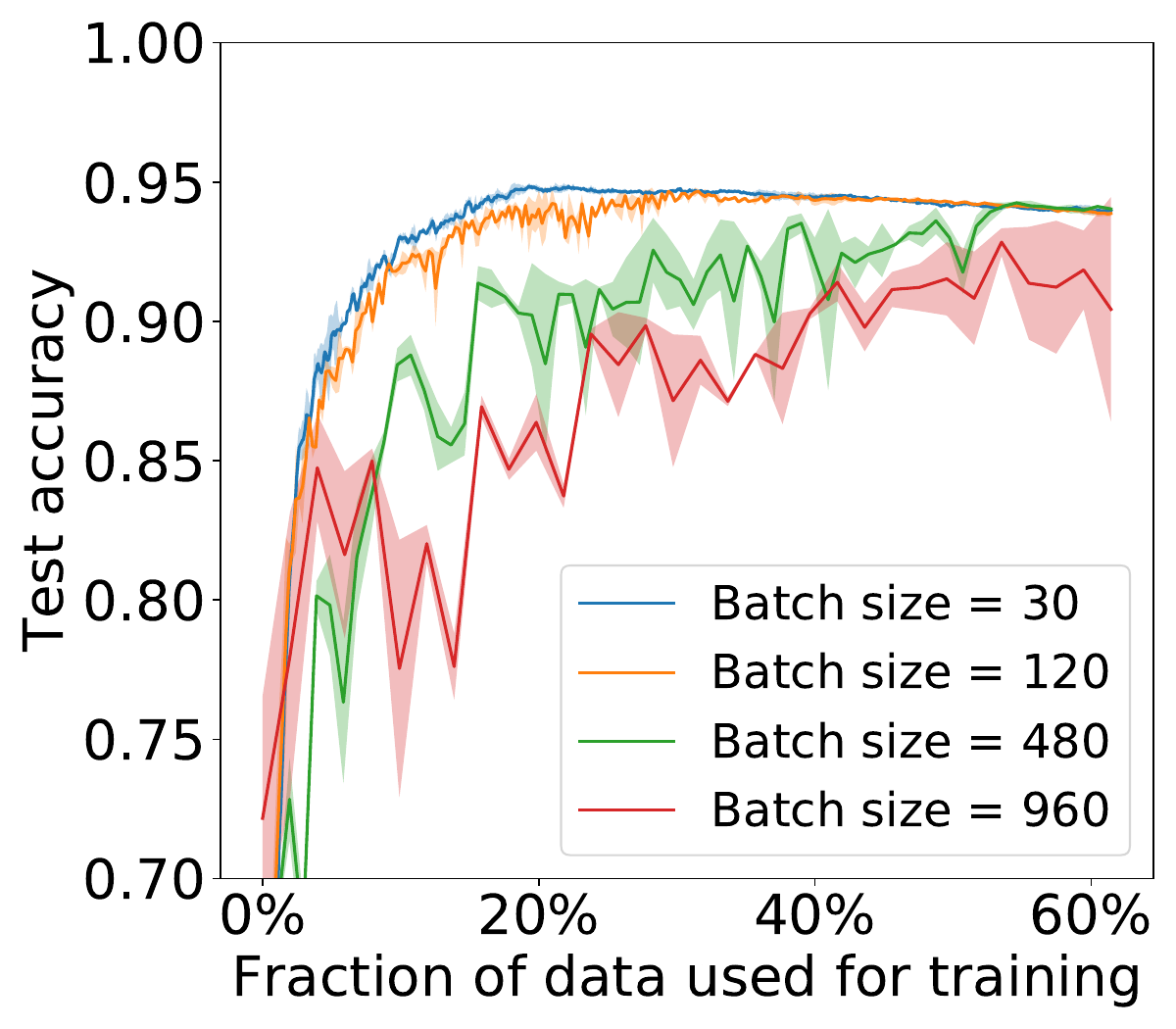}
}

\subfigure[Cumulative Fashion-MNIST.]{
    \includegraphics[width=0.22\textwidth]{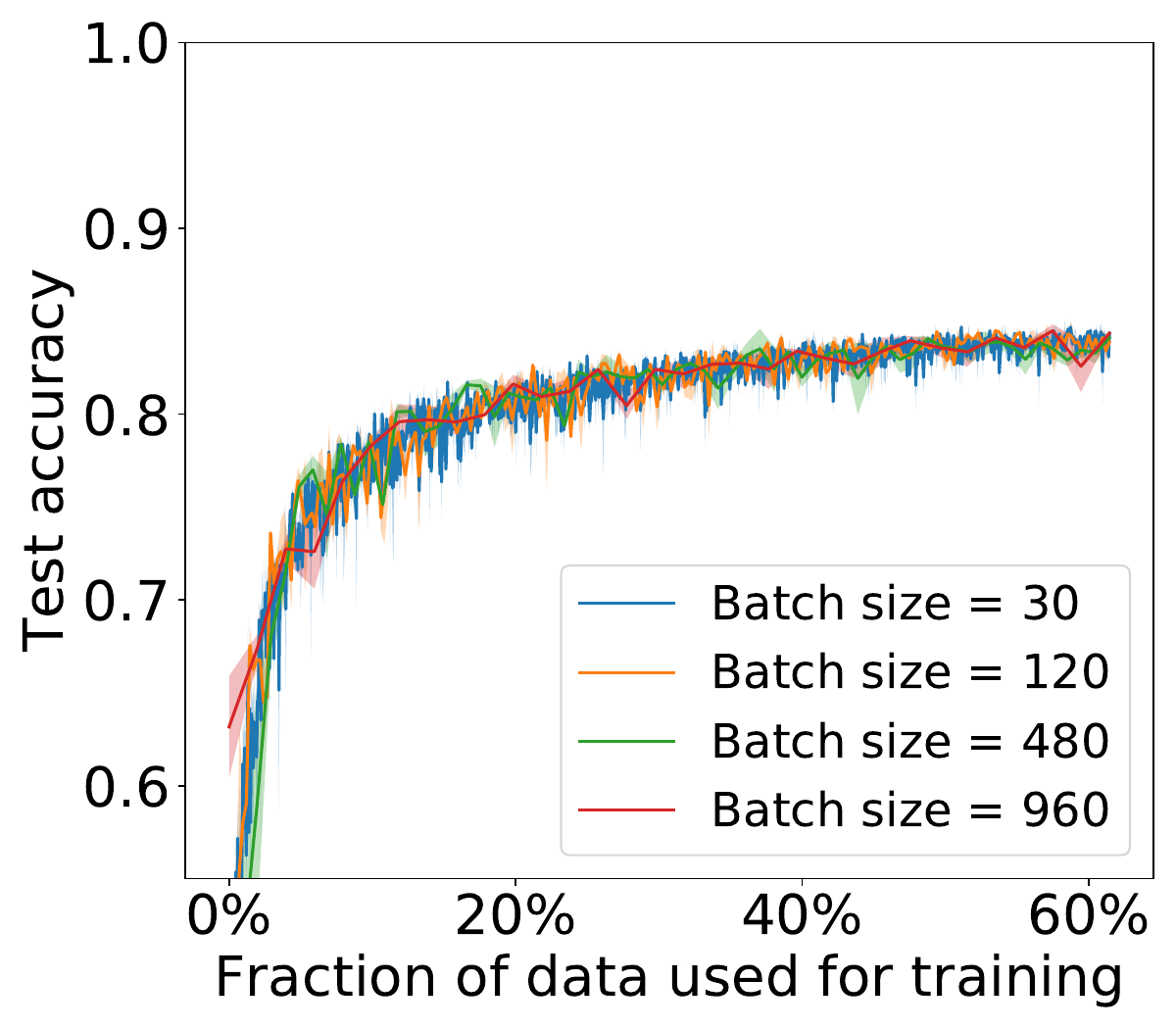}
}
\subfigure[Incremental Fashion-MNIST.]{
    \includegraphics[width=0.22\textwidth]{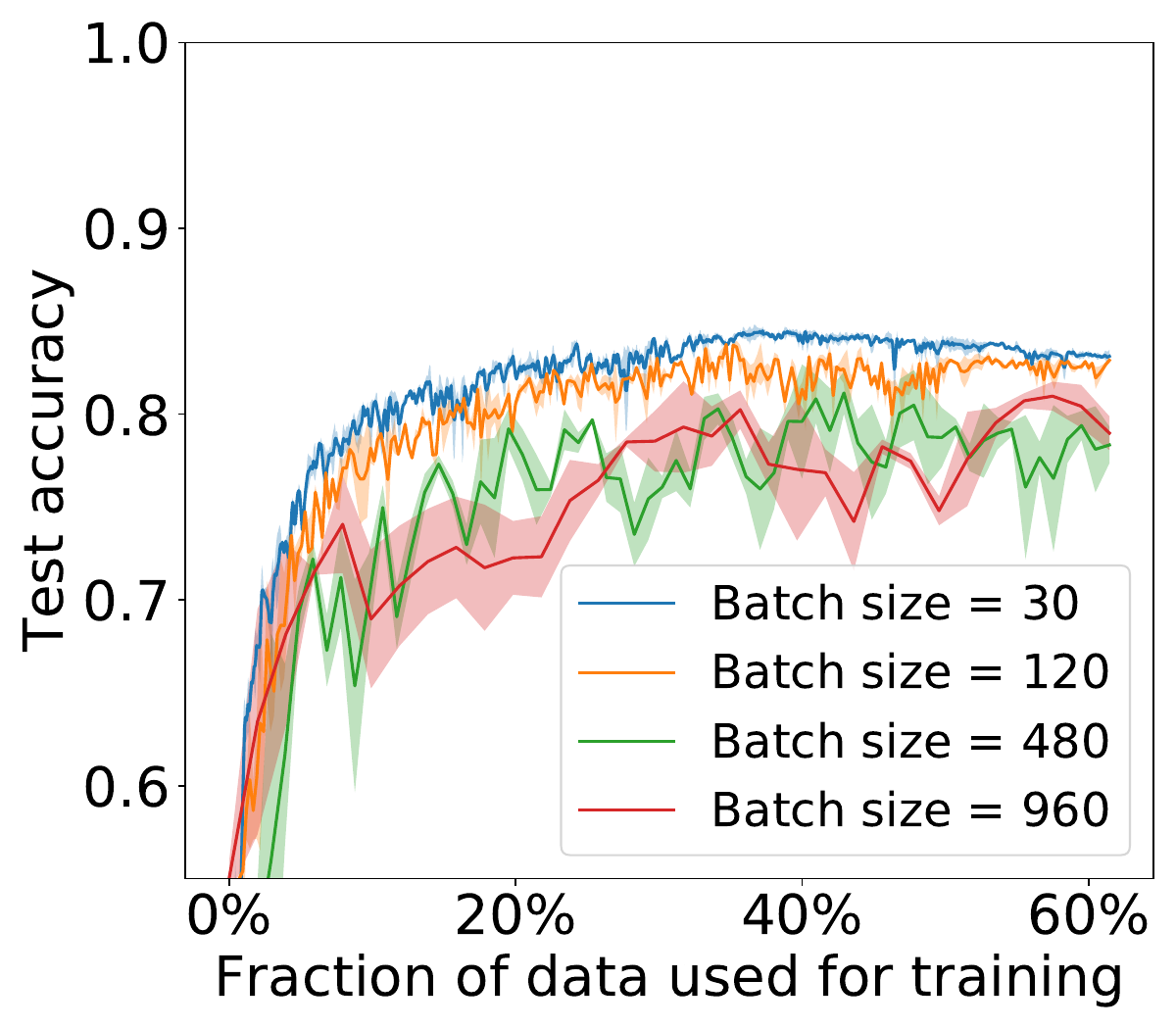}
}
\caption{The test accuracy for a number of different batch sizes using an $\mathrm{ANN1}_{100,1}$ network. The object labels are selected according to margin query strategy and the results are averaged over 2 runs.}
\label{fig:varying_batch_size:mnist:fashion}

\end{figure}

Figure \ref{fig:varying_batch_size:mnist:fashion} illustrates the test accuracy for different choices of $n_B$ for an $\mathrm{ANN1}_{100,1}$ network. We investigate both the incremental and cumulative modes for selecting object labels with the margin query strategy. The results are averaged over 2 runs.

The results show that the choice of the batch size can affect the performance of the network and query strategy in particular when incremental training is used. The cumulative training mode seems less sensitive to the choice of $n_B$.
The incremental training mode may be more affected by the choice of $n_B$ since the class assumed to be the most informative may change during the training process. A large $n_B$ may thus cause the network to train based on many objects wherein only one or few classes are represented. The cumulative training is not affected by a large batch size, presumably because it trains on all the available labeled data $L$. When the cumulative training mode selects a batch wherein only one or few classes are represented, the effect of the choice of the selected batch will be diluted by the objects labeled previously.

In most of the studies in this paper, either a batch size $n_B$ of 120 or a similar number such as 100 has been used. The results in Figure \ref{fig:varying_batch_size:mnist:fashion} show that different choices of $n_B$ for cumulative training yield consistent results and the exact choice might not matter so much.

\section{Informativeness Measures}

\paragraph*{Methodology}
\label{sec:background:informativeness_measures}

Different query strategies usually assign an informativeness measure to unlabeled data. In this study, we investigate how the informativeness measures relate to how quickly the performance (accuracy) of the network improves.
For example, at some point labeling new data may be expensive compared to the improvement in test accuracy.
We thus study if the training accuracy and informativeness measure may be used to evaluate the current state of the network, i.e., we formulate a stopping criterion for labeling (querying) without the need for a  dedicated test set.

It can be difficult to compare informativeness measures from different query strategies since they have different interpretations and ranges.
Thus, to determine the informativeness of different batches, we use the same informativeness measure, independent of how the batch is sampled. In this study we focus on the informativeness measure associated with random sampling $R$ and the cumulatively trained margin query strategy $M$. These choices are consistent with the results from Section \ref{sec:results_and_discussion:query_strategies_and_training_methods}.
The \textit{margin on random informativeness measure} (MOR) is obtained  by the mean margin informativeness measure $\MIM_B^M$ on a randomly selected batch $B$ with a network trained on the randomly selected data $L$.

\paragraph*{Results and discussion}
\label{sec:results_and_discussion:investigation_of_informativeness_measures}

\begin{figure*}[t]
    \centering

    \subfigure[MNIST.]{\includegraphics[width=0.28\textwidth]{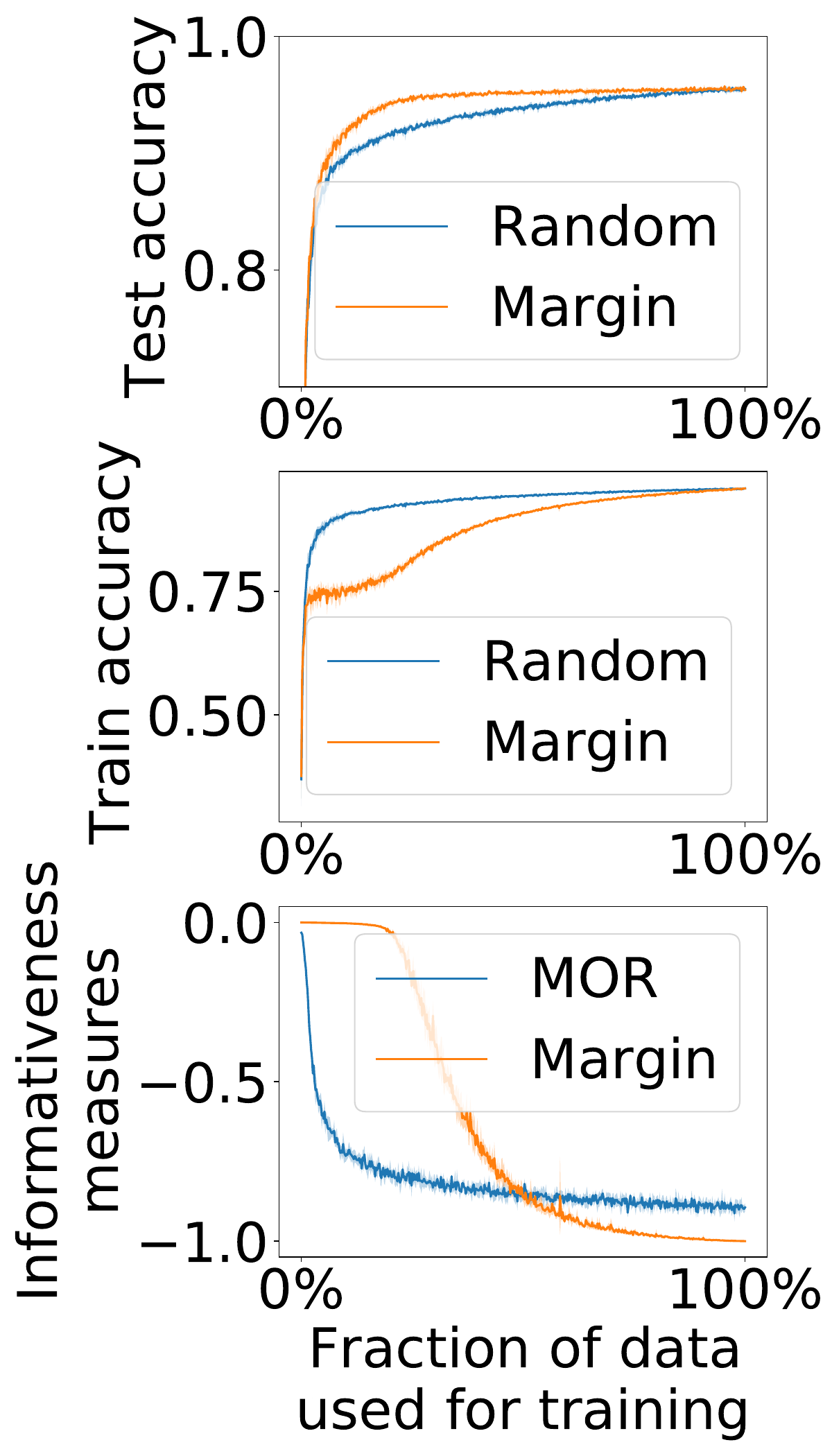}\label{fig:investigation_of_informativeness_measures:100:mnist} }
    \subfigure[Fashion-MNIST.]{\includegraphics[width=0.28\textwidth]{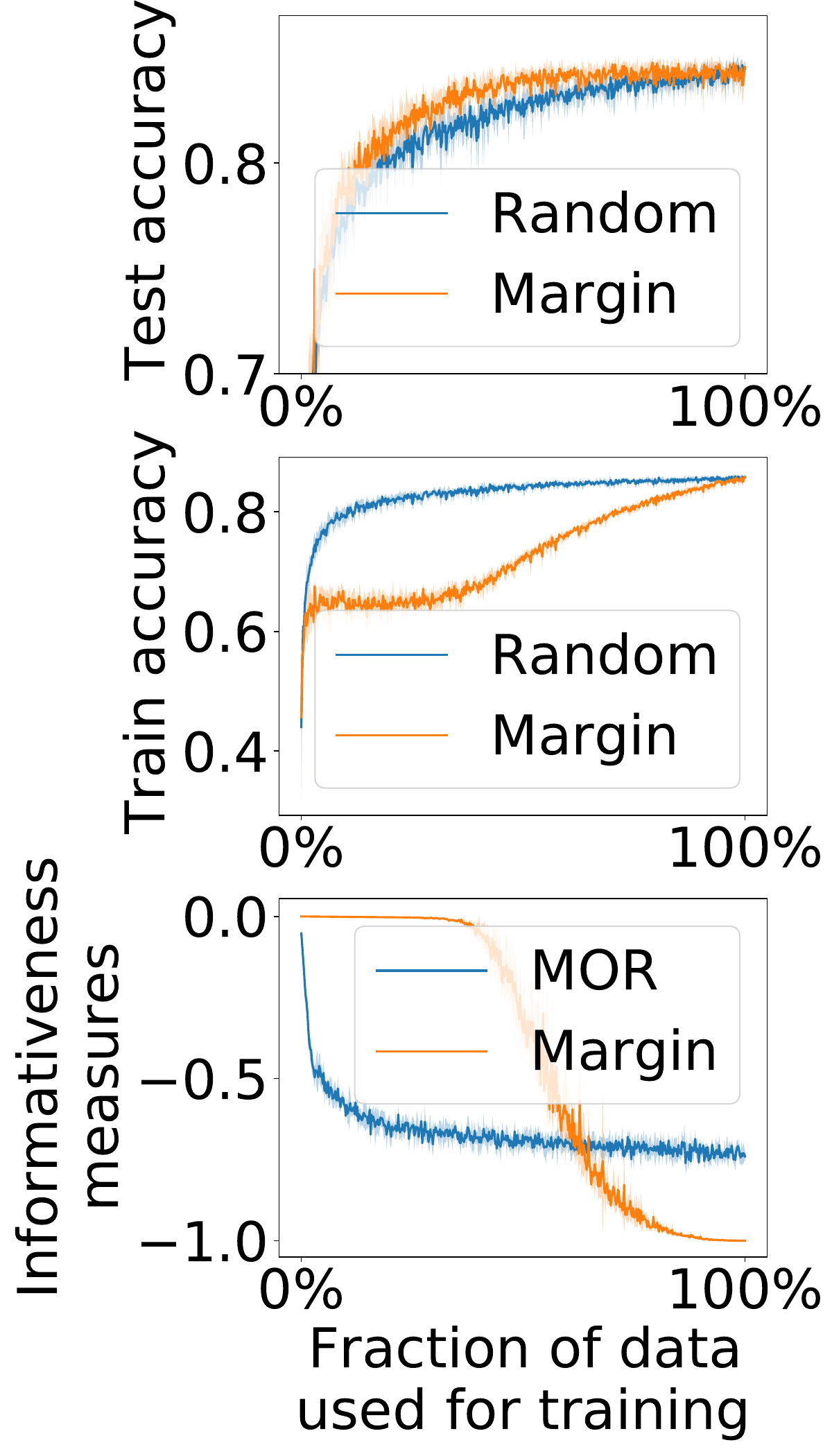}\label{fig:investigation_of_informativeness_measures:100:fashion} }
    \subfigure[CIFAR-10.]{\includegraphics[width=0.28\textwidth]{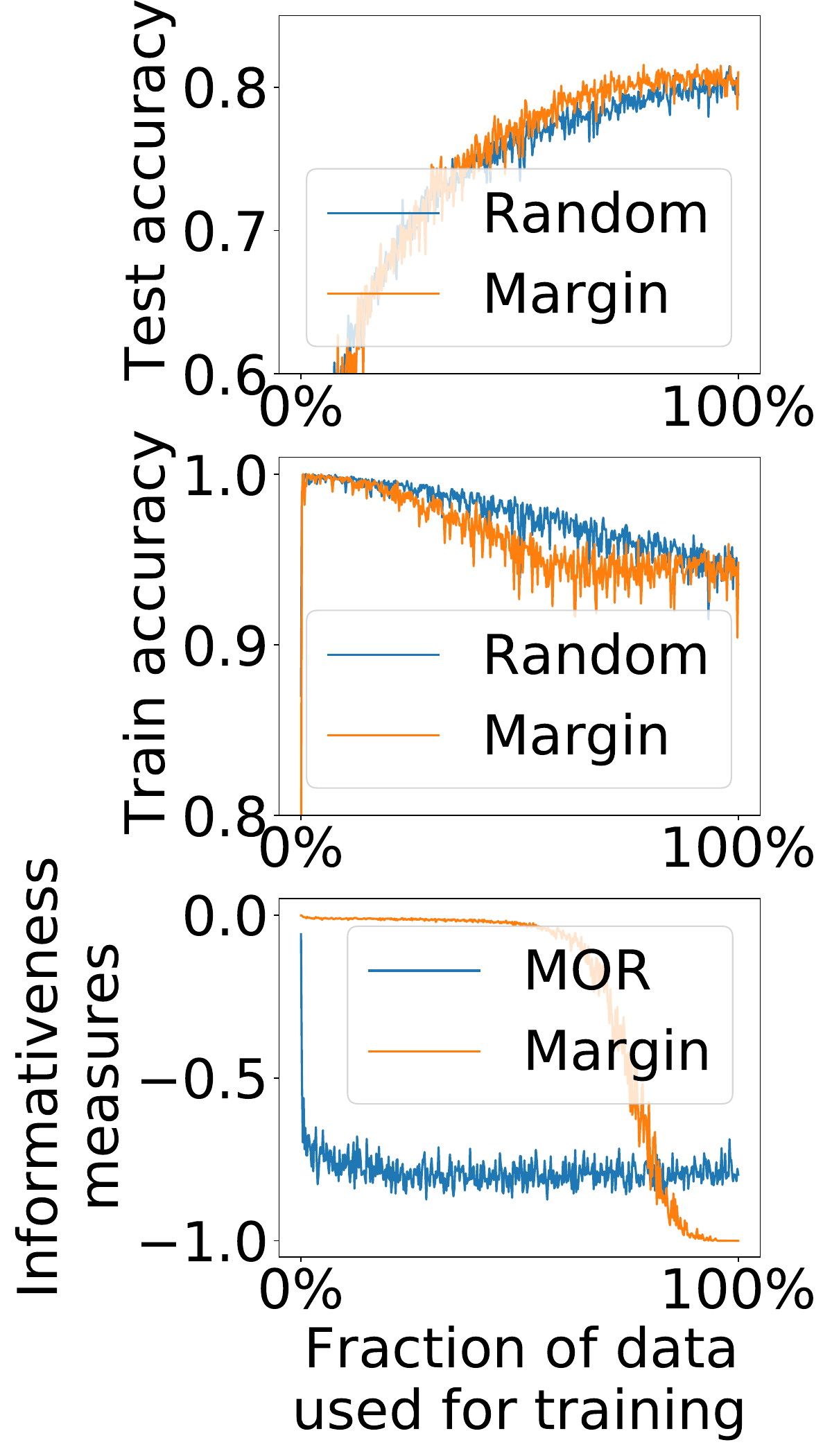}\label{fig:investigation_of_informativeness_measures:no_early_stopping:cifar10}}
    \caption{Test accuracy, training accuracy, and $\MIM^M_B$ measure for the query strategies $M$ and $R$.
    }
    \label{fig:investigation_of_informativeness_measures:mnist:fashion:cifar10}
\end{figure*}

Figure \ref{fig:investigation_of_informativeness_measures:mnist:fashion:cifar10} shows how the test accuracy, the train accuracy and $\MIM_B^M$ change while training with a query strategy. The results are obtained with cumulative training with a batch size of 120. The figure shows the results for the three data sets MNIST, Fashion-MNIST and CIFAR10, where we have used  $\mathrm{ANN1}_{100,1}$, $\mathrm{ANN1}_{100,1}$ and $\mathrm{CNN1}$ networks respectively.

We observe that the query strategy $M$ yields a higher test accuracy, a lower train accuracy, and the respective $\MIM^B_M$ is higher during the early steps of training, compared to the MOR measure that uses $R$.
Interestingly, with the margin query strategy, we sometimes obtain a higher test accuracy compared to its training accuracy.
It is usually expected that classifiers optimized on training sets have higher training accuracy than test accuracy on an equally distributed test set \cite[p.~31]{James2013}.
This suggests that the test and train sets are not equally distributed. Seeing a lower training accuracy than test accuracy could be interpreted as the margin query strategy actively selects ``difficult'' objects for labeling and training, and thus the train accuracy is lower.
However, the data that the margin query strategy selects is seemingly better for describing the distribution of the test set, indicated by the fact that it yields a higher test accuracy compared to the $R$ strategy. Thus, to ensure a maximal test accuracy using as few objects as possible, we want the $\MIM^M_B$ to be high. Furthermore, $\MIM^M_B$ can be linked to learnability of the network in the sense that a high $\MIM^M_B$ yields a good improvement in the test accuracy. This means that a decreasing $\MIM^M_B$ can be interpreted as either: (a) an indication that we are not gaining enough accuracy by querying the oracle for labels in the training set and should stop training the network, or,
(b) more unlabeled data should be acquired in order to improve accuracy. These observations can be useful in the cases no test set is available to measure the test accuracy.

We also notice the correlation between the increase in train accuracy and  the decrease in $\MIM^M_B$. We may interpret it in two ways: (a) The network learns to characterize the data, which makes the network more certain in its predictions of the objects in $B$, thus  $\MIM^M_B$ is decreasing. (b) The unlabeled data set $U$ is depleted of (informative) objects close to the decision boundaries, since they have a higher $I_i^M$ and are thus found earlier in the training. The objects that are left are further away from the decision boundaries, and hence, have a lower $I_i^M$.
We study this behavior in more detail in Section \ref{sec:data_split_and_sampling}.

The training accuracy on the CIFAR-10 data set evolves differently compared to MNIST and Fashion-MNIST, as seen in Figure \ref{fig:investigation_of_informativeness_measures:no_early_stopping:cifar10}. The training accuracy starts from a high value and decreases as more data is labeled and used for training, independent of the query strategy used. This is possibly due to  training the network with 50 epochs which implies the (small) training set at the beginning can be almost perfectly described by the weights (and the relatively high capacity). We see a slightly better performance in test accuracy when using $M$ compared to $R$ after around 50\% of the data has been used for training. The observations regarding the $\MIM^M_B$ are consistent for all the examined data sets.

\section{Data Split and Sampling}
\label{sec:data_split_and_sampling}

\paragraph*{Methodology}
Evaluating query strategies on the entire unlabeled data set $U$ can be computationally expensive. On the other hand, it might be possible that more unlabeled data is available after active learning has begun.
Therefore, in this section, we study the setting wherein the full $U$ is not accessible for query strategies.
In particular, we first
investigate sampling (splitting) unlabeled data for active learning. We then  study why the informativeness measure decreases when using the margin query strategy. As illustrated in Figure \ref{fig:investigation_of_informativeness_measures:mnist:fashion:cifar10},
during the later steps of the training, when almost all the data has been labeled, $\MIM^M_B$ decreases.
There are two hypotheses to explain the decrease of $\MIM_B^M$:
(1) the network learns enough, such that it reaches its best performance and no further improvement is possible;
(2) when selecting data with a query strategy, the most informative objects are labeled first, leaving uninformative objects in $U$. This means that the decrease of $\MIM_B^M$ is due to exhaustion of informative data.
It is likely that both hypotheses contribute to the decrease of $\MIM_B^M$, but the extent might be different.

If hypothesis (2) is true, then it would be useful to ensure that $U$ is as large as possible and extend it when the informativeness measure decreases.
To understand this, we split the
unlabeled data set into $d$ subsets, i.e. $U_1, \dots, U_d$, which are independently and identically distributed. The margin query strategy queries a fraction $p$ of objects from $U = U_1$ to be labeled actively and moved to $L$. We then use the next unlabeled subset, i.e. $U \leftarrow U \cup U_2$. We perform this procedure sequentially for all the $d$ subsets. In particular, we perform two types of experiments with the following parameters on both the MNIST and Fashion-MNIST data sets.
{\centering
    \begin{tabular}{ccll}
        $d$ & $p$ & batch size & number of samples in each subset \\ \hline
        2 & 80\% & 120 & 30000 \\
        5 & 90\% & 200 & 12000
    \end{tabular}
}
In this setup, the increase in $\MIM_B^M$ upon adding a new subset $U_j$ is an indication of depletion of informative data in the current $U$.

\begin{figure*}[t]
    \centering
    \subfigure[MNIST, 2 splits.]{\includegraphics[width=0.33\textwidth]{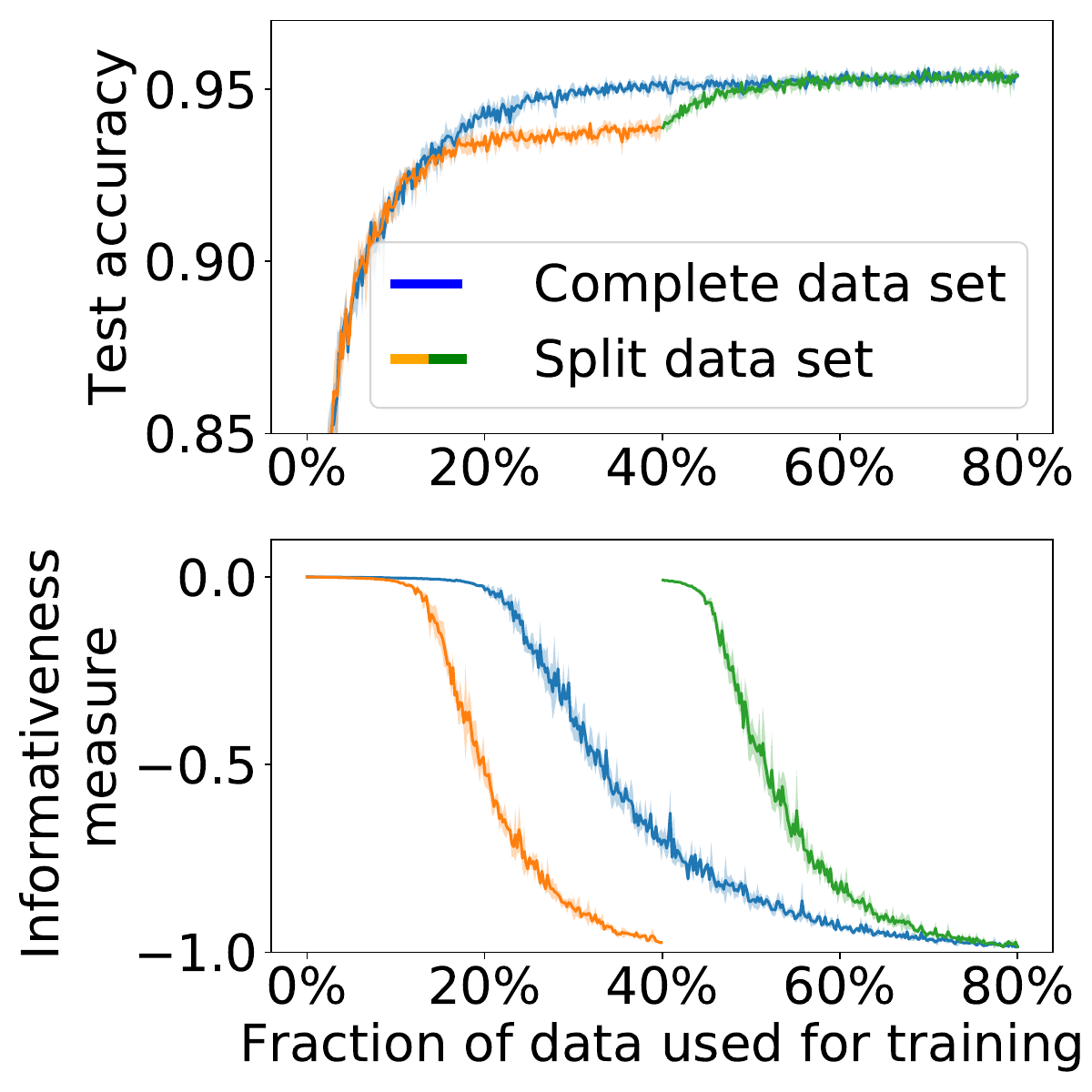} \label{fig:investigation_of_informativeness_measures:high_margins:mnist:2split}}
    \subfigure[Fashion-MNIST, 2 splits.]{\includegraphics[width=0.33\textwidth]{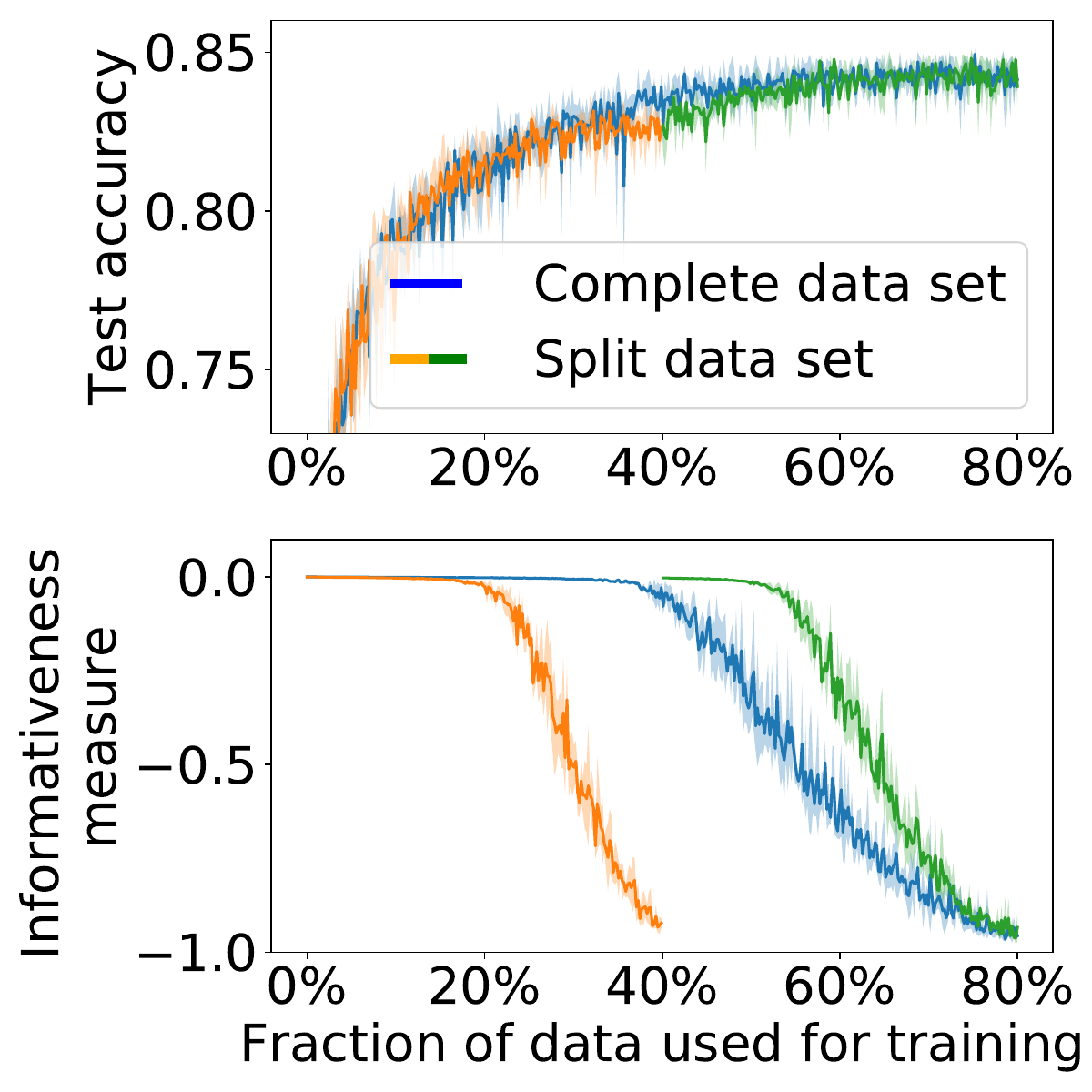} \label{fig:investigation_of_informativeness_measures:high_margins:fashion:2split}}
    \\
    \subfigure[MNIST, 5 splits.]{\includegraphics[width=0.33\textwidth]{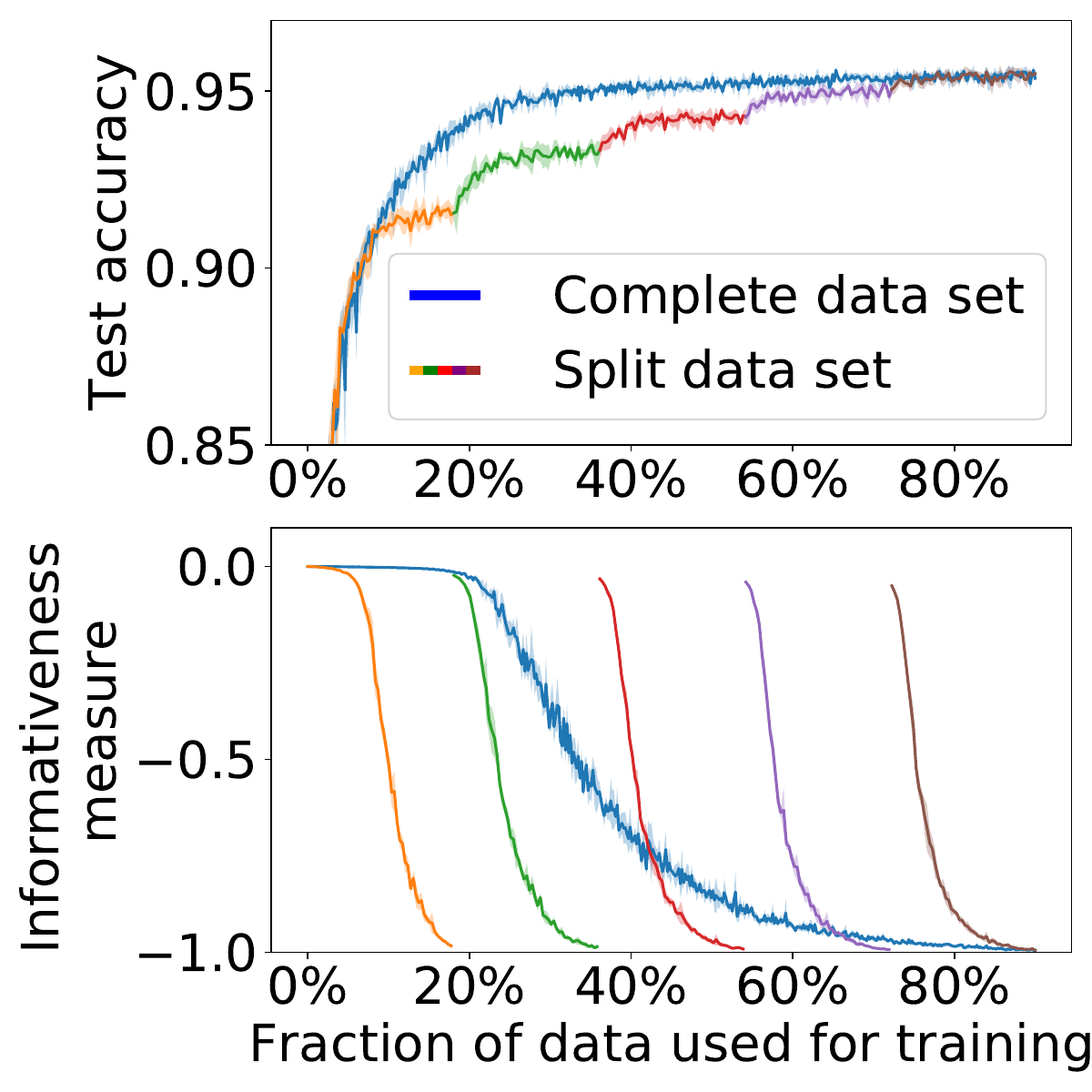} \label{fig:investigation_of_informativeness_measures:high_margins:mnist:5split}}
    \subfigure[Fashion-MNIST, 5 splits.]{\includegraphics[width=0.33\textwidth]{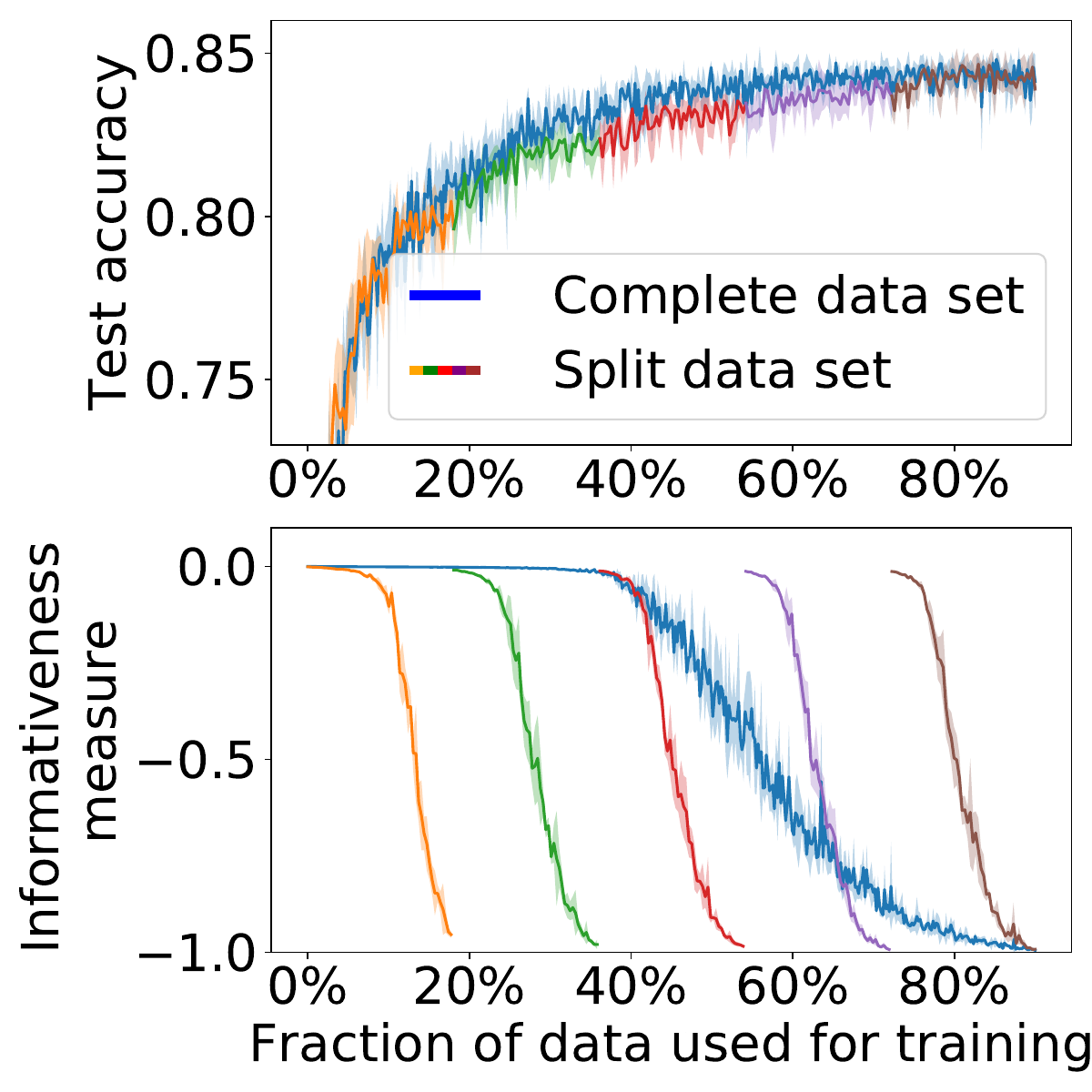} \label{fig:investigation_of_informativeness_measures:high_margins:fashion:5split}}
    \caption{
    The effect of data splits on the informativeness measure and test accuracy when new unlabeled subsets are added during active learning. 
    }
    \label{fig:investigation_of_informativeness_measures:high_margins:mnist:fashion:split}
\end{figure*}

\paragraph*{Results and discussion}
Figure \ref{fig:investigation_of_informativeness_measures:high_margins:mnist:fashion:split} shows the test accuracy and mean informativeness measure $\MIM^M_B$ for different experimental settings. The result corresponding to ``Complete data set'' is provided for a comparison with having access to all unlabeled data from the beginning.
%
In Figure \ref{fig:investigation_of_informativeness_measures:high_margins:mnist:fashion:split}, we observe sharp phase transitions in  $\MIM^M_B$ that correspond to drastic increase of $\MIM^M_B$ upon adding every new subset to $U$. Even with the last subset in Figures \ref{fig:investigation_of_informativeness_measures:high_margins:mnist:5split} and \ref{fig:investigation_of_informativeness_measures:high_margins:fashion:5split}, where the model is already trained using 43,200 objects, $\MIM^M_B$ instantly increases from -1 to 0.
This observation suggests that the decrease of $\MIM^M_B$  seen in Figure \ref{fig:investigation_of_informativeness_measures:mnist:fashion:cifar10} is mainly because the unlabeled data set does not contain informative objects anymore.

We also observe the increase of the test accuracy as new subsets are added to $U$, as shown in Figures \ref{fig:investigation_of_informativeness_measures:high_margins:mnist:2split} and \ref{fig:investigation_of_informativeness_measures:high_margins:mnist:5split}. This is in particular more significant when $\MIM^M_B$ is high. This reinforces the idea that the rate of increase in test accuracy is related to the $\MIM^M_B$, as discussed in Section \ref{sec:background:informativeness_measures}. The more informative the data is, the faster the increase in accuracy occurs.
We also observe in Figure \ref{fig:investigation_of_informativeness_measures:high_margins:mnist:fashion:split} that it would be preferable to have access to all the unlabeled data from the beginning compared to gathering them during the active learning process. The test accuracy when all the data is available is always higher. 

The phase where $\MIM^M_B$ is close to 0 is the longest when $U_1$ is added. It becomes shorter for each new subset added as shown in Figure \ref{fig:investigation_of_informativeness_measures:high_margins:mnist:fashion:split}. As the network gets more trained, the number of batches with approximately zero $\MIM^M_B$ decreases.
Therefore, from the results illustrated in Figure \ref{fig:investigation_of_informativeness_measures:high_margins:mnist:fashion:split} we conclude that $\MIM^M_B$ is high as long as there are informative objects in the unlabeled set $U$. This is consistent with the observation in Figure \ref{fig:investigation_of_informativeness_measures:mnist:fashion:cifar10} showing that the training accuracy increases faster when $\MIM^M_B$ is high.

\section{Random Start}

\paragraph*{Methodology}
\label{sec:background:random_start}
When we start active learning, at the beginning, the network might not be trained properly (e.g., entropy in Figure \ref{fig:query_strategies:medium:fashion}). Thus, the model predictions might be inaccurate. This means that the first batch selected by a query strategy might be biased towards a certain class, in particular if the network is initialized randomly.
To investigate if such a bias exists, we examine the number of occurrences of the different classes that the query strategies select. We do this by initializing an $\mathrm{ANN1}_{100,1}$ network
and letting it select 1000 images using a query strategy. This procedure is repeated 200 times where we examine the $M$ and $E$ query strategies and compare them to $R$.

Our hypothesis is that the query strategies might be biased towards dark images. We explore this idea by generating images of size $28 \times 28$ pixels containing uniformly random noise. A certain percentage $\alpha$ of the pixels, selected randomly, are turned black. If the images with $\alpha$ close to 1 have higher $I^A_i$, we conclude that the query strategy $A$ is biased. We repeat this 100 times using a randomly initialized $\mathrm{ANN1}_{100,1}$ network.

We then argue that the bias effect can be mitigated by sampling with $R$ initially. We refer to this as a \textit{random start}. The \textit{random start size} is the number of objects that are selected randomly for labeling. We perform the experiments with different random start sizes such as 2000 and with a cumulatively trained $\mathrm{ANN1}_{100,1}$ network to see if the random start yields any performance benefits.

\paragraph*{Results and discussion}

\begin{figure*}[th]
    \centering
    \subfigure[Random.]{\includegraphics[width = 0.25\textwidth]{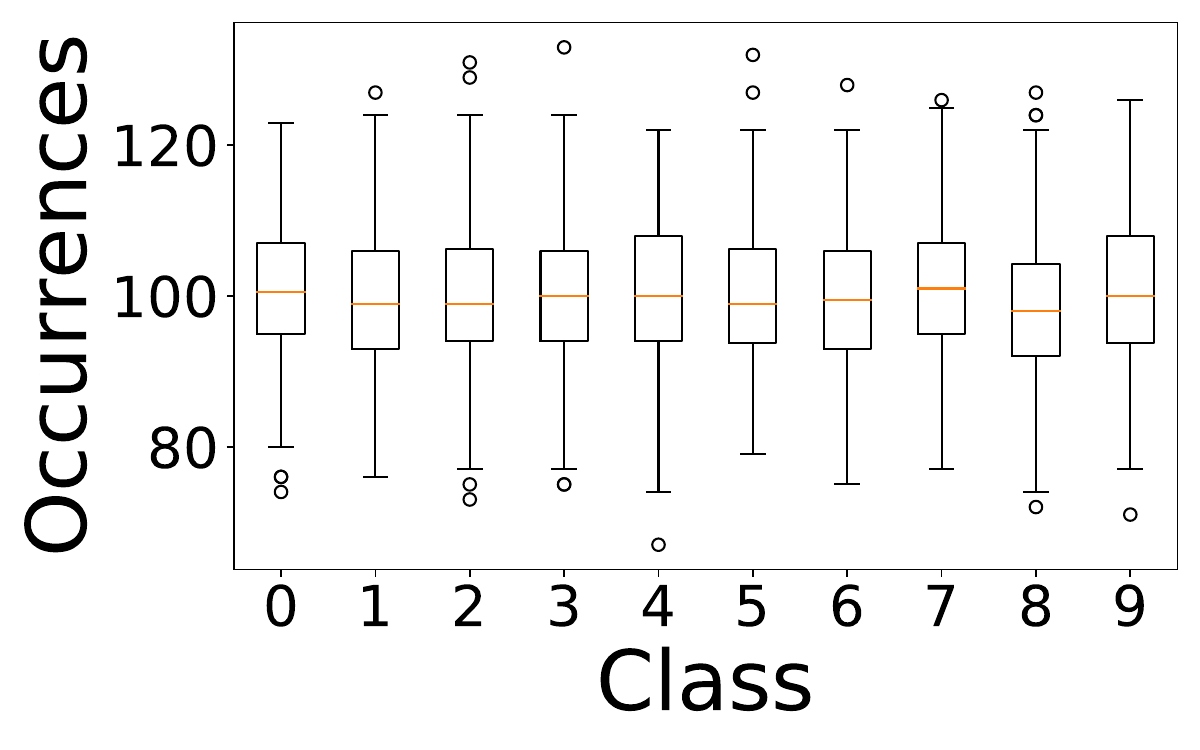} \label{fig:random_start:random:fashion}}
    \subfigure[Margin.]{\includegraphics[width = 0.25\textwidth]{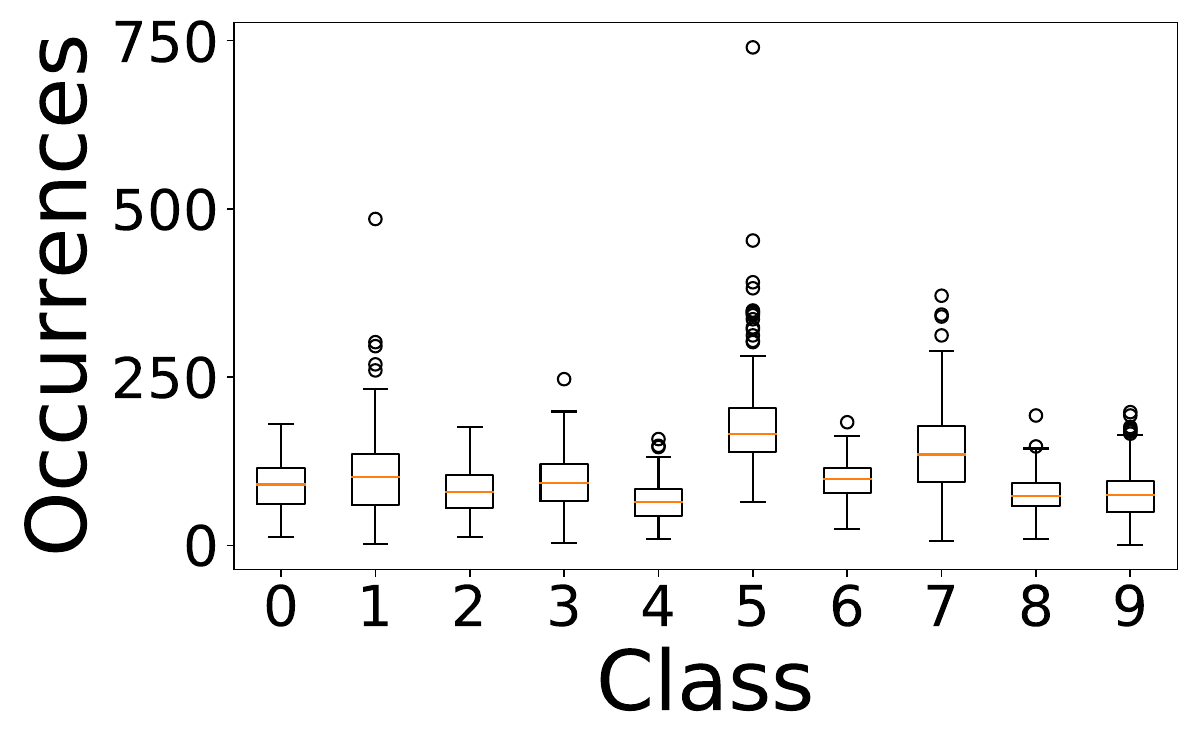} \label{fig:random_start:margin:fashion}}
    \subfigure[Entropy.]{\includegraphics[width = 0.25\textwidth]{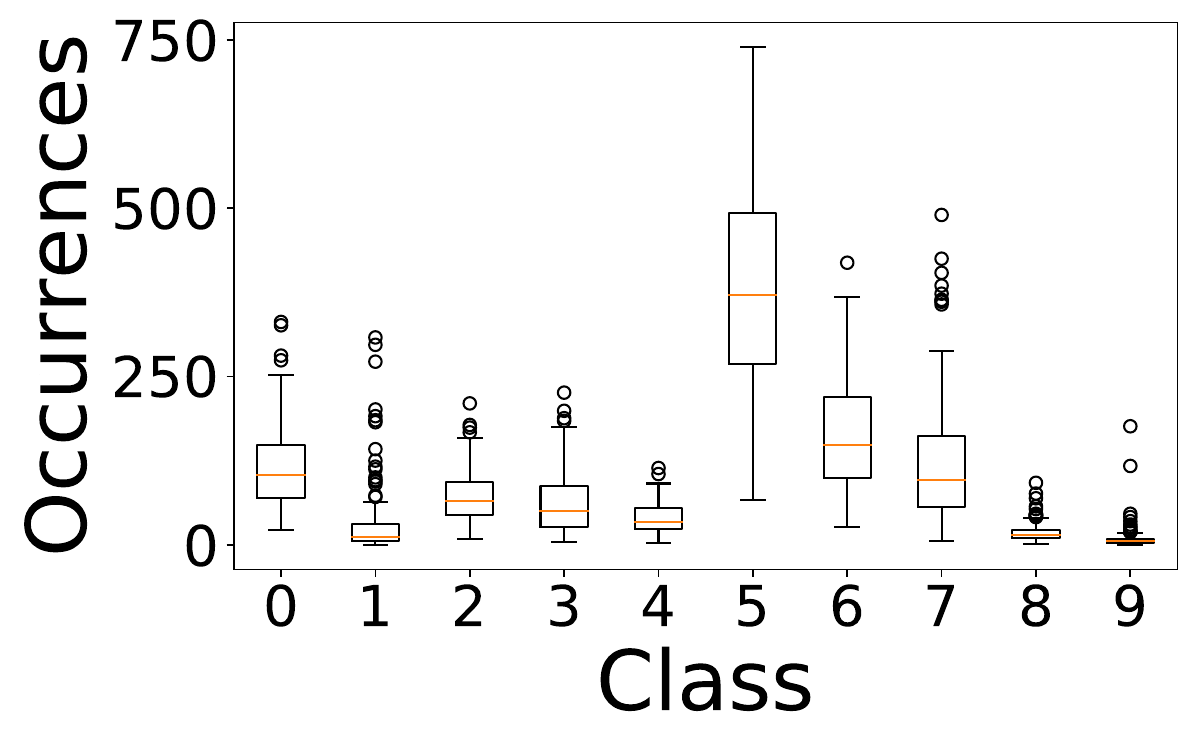} \label{fig:random_start:entropy:fashion}}
\\
    \subfigure[Random.]{\includegraphics[width=0.24\textwidth]{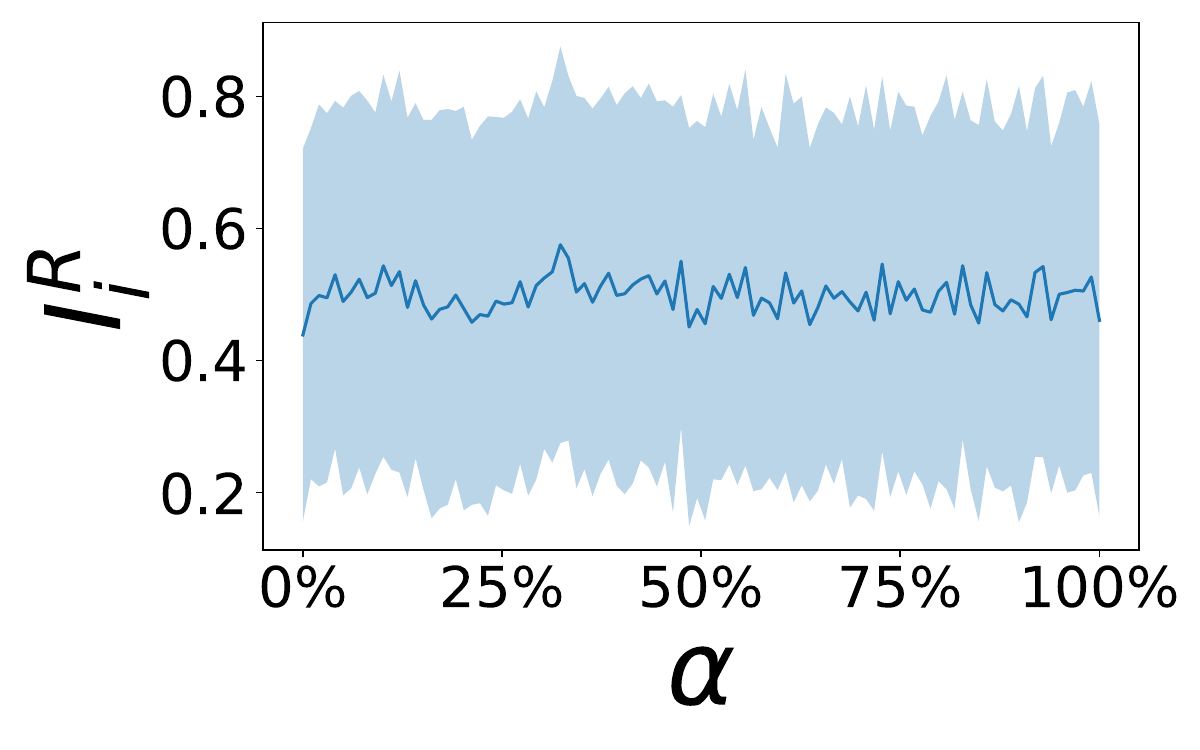} \label{fig:random_start_randombias}}
    \subfigure[Margin.]{\includegraphics[width=0.24\textwidth]{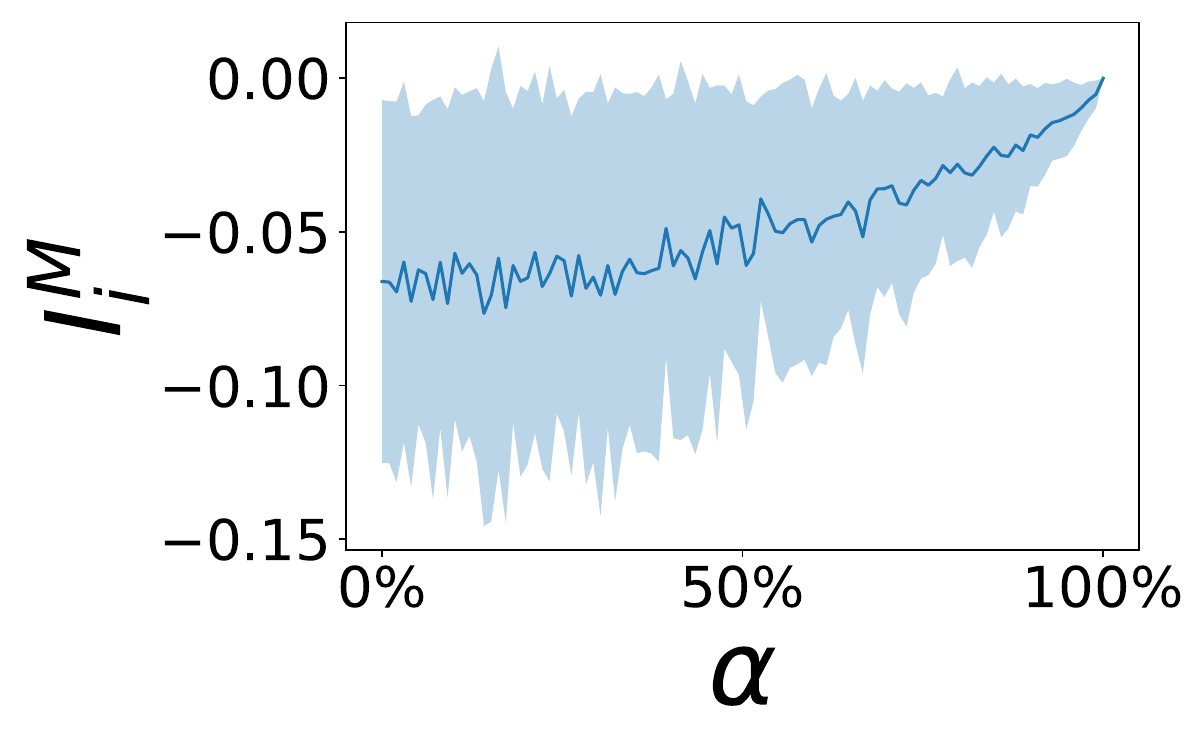} \label{fig:random_start_marginbias}}
    \subfigure[Entropy.]{\includegraphics[width=0.24\textwidth]{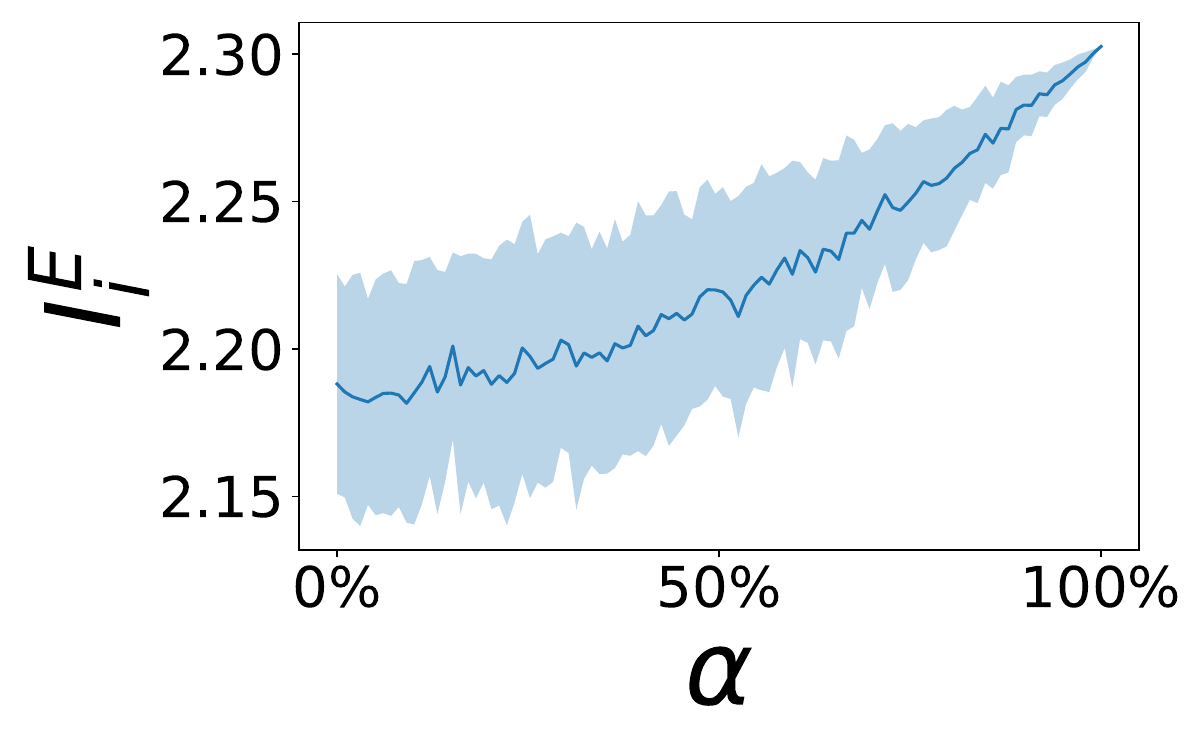}\label{fig:random_start_entropybias}}
    \caption{
Sampled class distribution obtained from different query strategies and a randomly initialized network on  Fashion-MNIST.
    }
    \label{fig:random_start:all:fashion}\label{fig:random_start:query_strategies_distribution:mnist:fashion}
\end{figure*}

Figure \ref{fig:random_start:all:fashion} (first row) shows the box-plots of the class occurrences for the Fashion-MNIST data set with the $M$, $E$ and $R$ query strategies.
The class distribution of the objects selected by $R$ is consistent with the distribution of the entire data set.
The query strategies $M$ and $E$ select objects in a way that the class 5 tends to be over-represented, corresponding to the images of \textit{Sandals}. This observation is even more obvious when using $E$.
We thus conclude that the query strategies might be biased in their selection when performing on an untrained network.

Figure \ref{fig:random_start:query_strategies_distribution:mnist:fashion} (second row) shows how $I^A_i$ changes depending on $\alpha$, averaged over 100 runs. $I^A_i$ increases with $\alpha$ for  $M$ and $E$ strategies. $M$ shows a relatively large variance compared to $E$, which possibly explains why its class distribution is closer to uniform in Figure \ref{fig:random_start_marginbias}.
Intuitively, an image of a ``Sandal'' should be darker compared to the other classes, such as ``T-shirt'' or ``Coat'' in the Fashion-MNIST data set, due to its smaller surface area and the background being dark. Figures \ref{fig:random_start_marginbias}  and \ref{fig:random_start_entropybias} thus explain why class 5 is over-represented in the first batch.
We guess the preference for darker images is mostly due to the combination of the initialization and the specific form of the query strategies.

\begin{figure}[t]
    \centering
    \hspace{-4mm}
    \subfigure[Random start.]{\includegraphics[width = 0.25\textwidth]{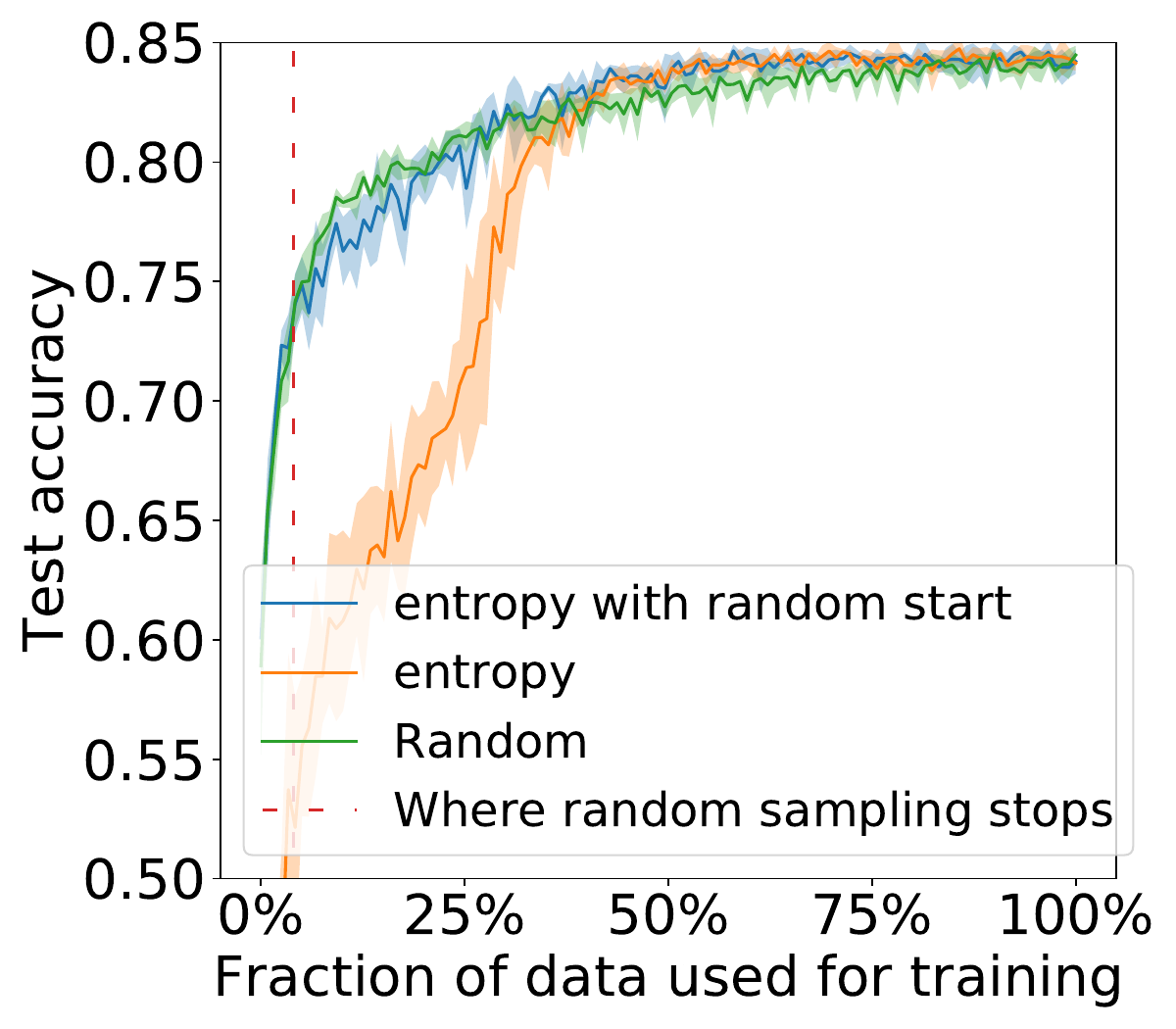} \label{fig:random_start:prolonging_effect_random_start_all_data:entropy:fashion}}
    \hspace{-3mm}
    \subfigure[Accuracy estimates.]{\includegraphics[width = 0.25\textwidth]{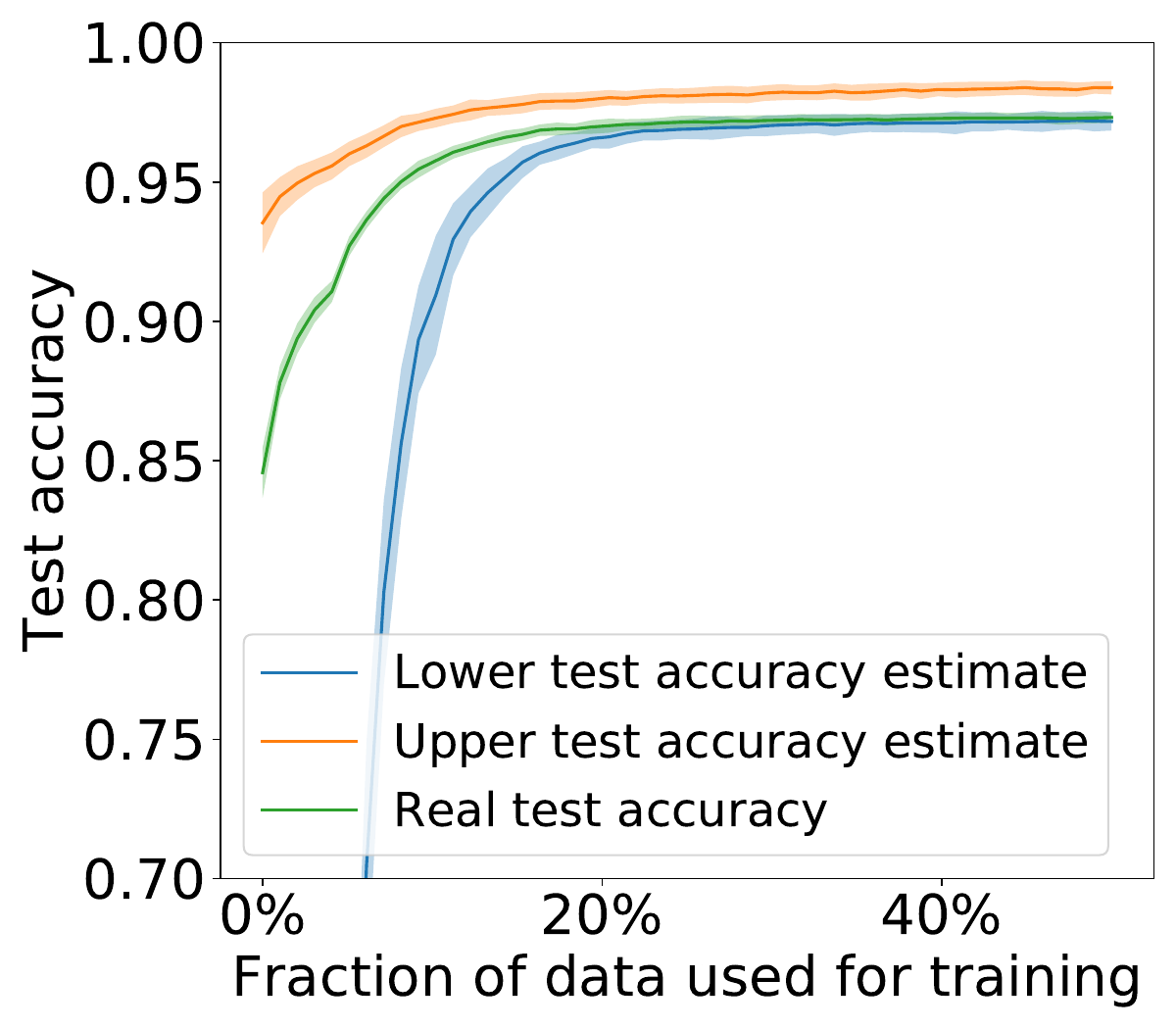} \label{fig:random_start:prolonging_effect_random_start_all_data:margin:fashion}}

    \caption{Test accuracy improvement with random start (a) and estimates of test accuracy (b) when using random start.}
    \label{fig:random_start:prolonging_effect_random_start_all_data:fashion}
\end{figure}

Figure \ref{fig:random_start:prolonging_effect_random_start_all_data:entropy:fashion} illustrates  the test accuracy with a random start of size 2000 for the entropy query strategy, which yields a significant performance improvement. 
With a smaller random start size of 200 we do not observe a significant improvement. The size is insufficient to obtain a reliable initial training of the network in order to mitigate the bias.

Additionally, as a side study,  we investigate the random start to compute heuristic empirical lower and upper bounds on the test accuracy.
This can in particular be useful when a separate labeled test data set is not available for evaluations. We consider the following
test and training sets for this purpose.
\begin{enumerate}
\item $L_t$: contains the random start objects and is used as a test set.
\item $\underline{L}$: contains all labeled objects excluding the random start, i.e.,  only the objects  labeled via the margin query strategy.
\item $L= L_t \cup \underline{L}$: includes all labeled data.
\end{enumerate}.
We obtain the empirical lower bound on test accuracy by training the model on $\underline{L}$ and evaluating on $L_t$. We obtain the empirical upper bound by training on $L$ and evaluating on $L_t$, which means that the test set is a subset of the training set. Figure \ref{fig:random_start:prolonging_effect_random_start_all_data:margin:fashion} shows that the test accuracy of an $\mathrm{ANN1}_{100,5}$ network on MNIST lies between the heuristic empirical lower and upper bounds, using a random start size of 3000.

\begin{figure}
    \centering
    \includegraphics[width = 0.28\textwidth]{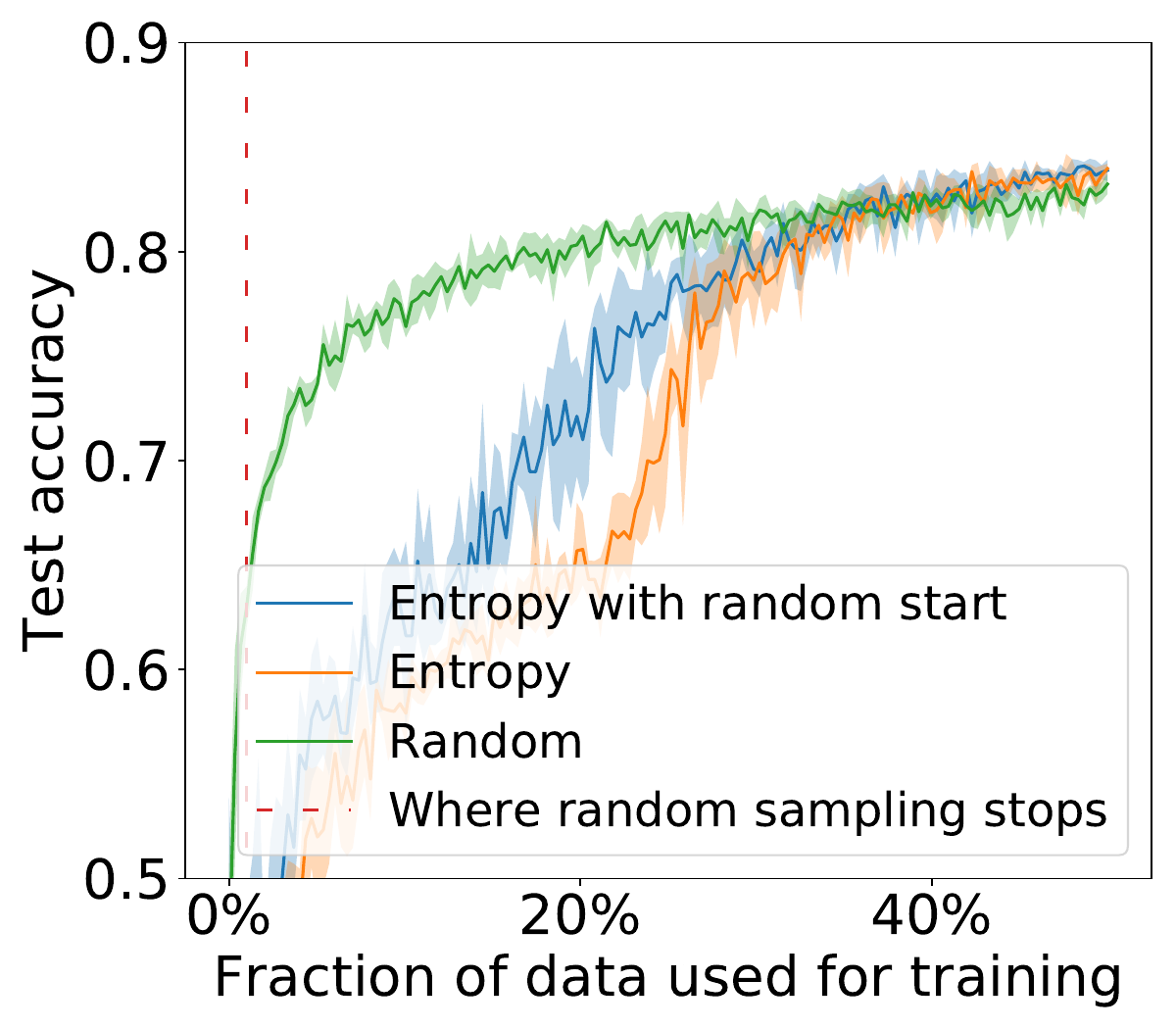}
    \caption{Entropy, random start size 200, for Fashion MNIST.}
    \label{fig:Entropyrandomstartsize200}
\end{figure}

Figure \ref{fig:Entropyrandomstartsize200} shows a random start of  size 200. We observe that although this random start improves the test accuracy, the used random start size is not sufficiently large to yield good performance. Thus, the random start size must be large enough to be effective.

\begin{figure}[thb!]
    \subfigure[Random start size 3000, $\mathrm{ANN1}_{100,1}$.]{
        \includegraphics[width=0.22\textwidth]{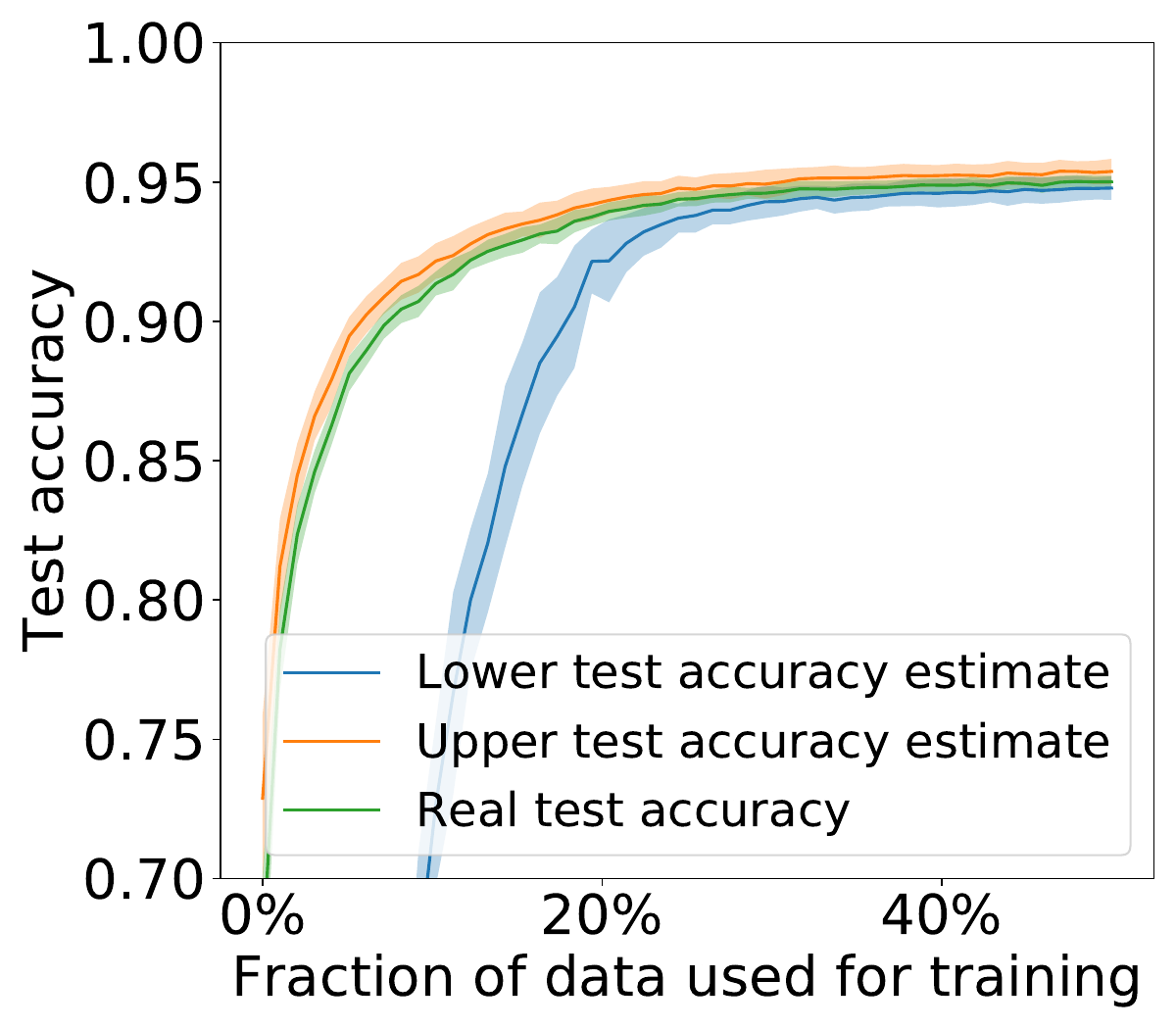}
        \label{fig:estimates_of_the_test_accuracy_475_25_1:mnist}
    }
    \hfill
    \subfigure[Random start size 3000, $\mathrm{ANN1}_{100,5}$.]{
        \includegraphics[width=0.22\textwidth]{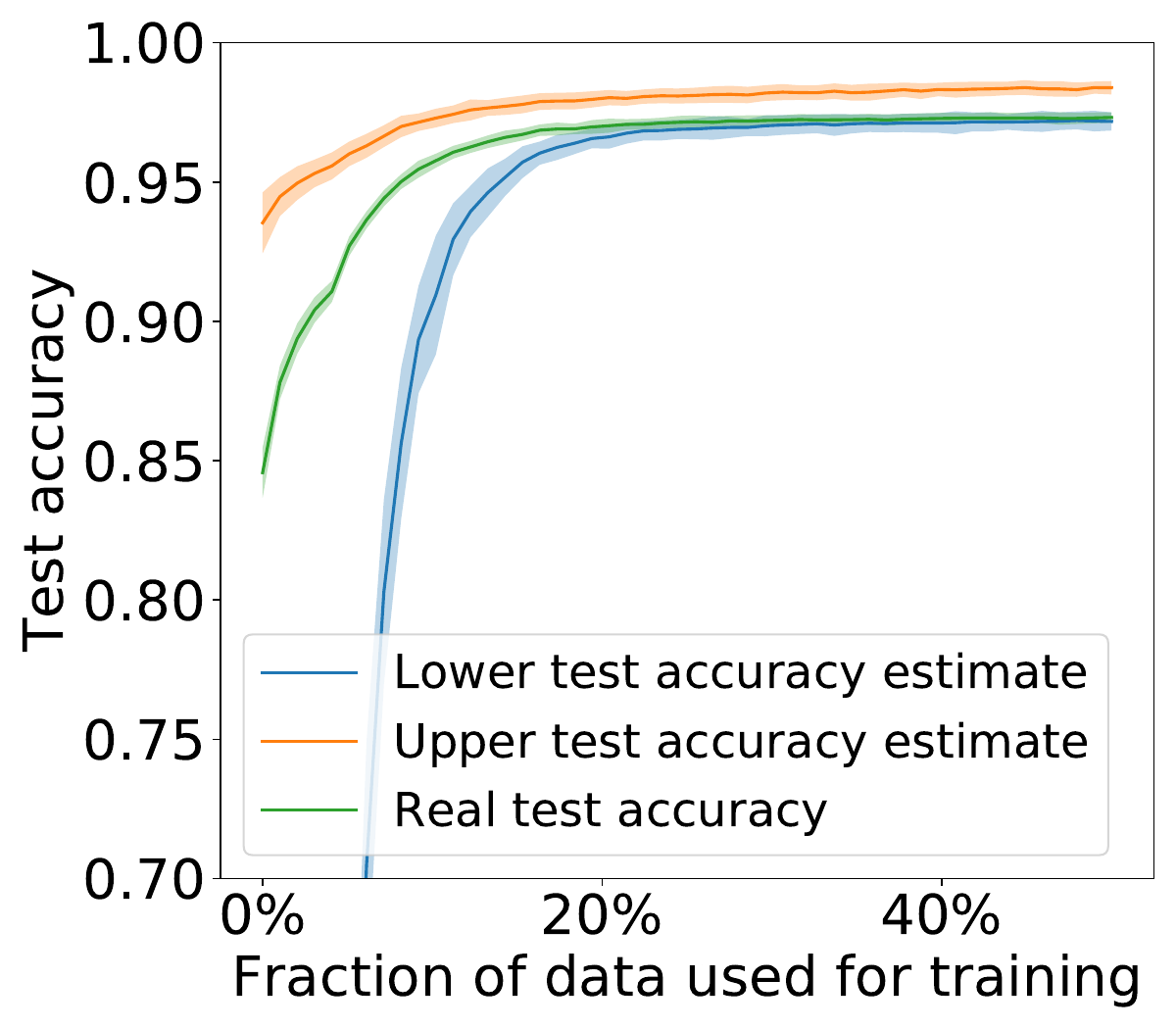}
        \label{fig:estimates_of_the_test_accuracy_475_25_5:mnist}
    }

    \subfigure[Random start size 600, $\mathrm{ANN1}_{100,1}$.]{
        \includegraphics[width=0.22\textwidth]{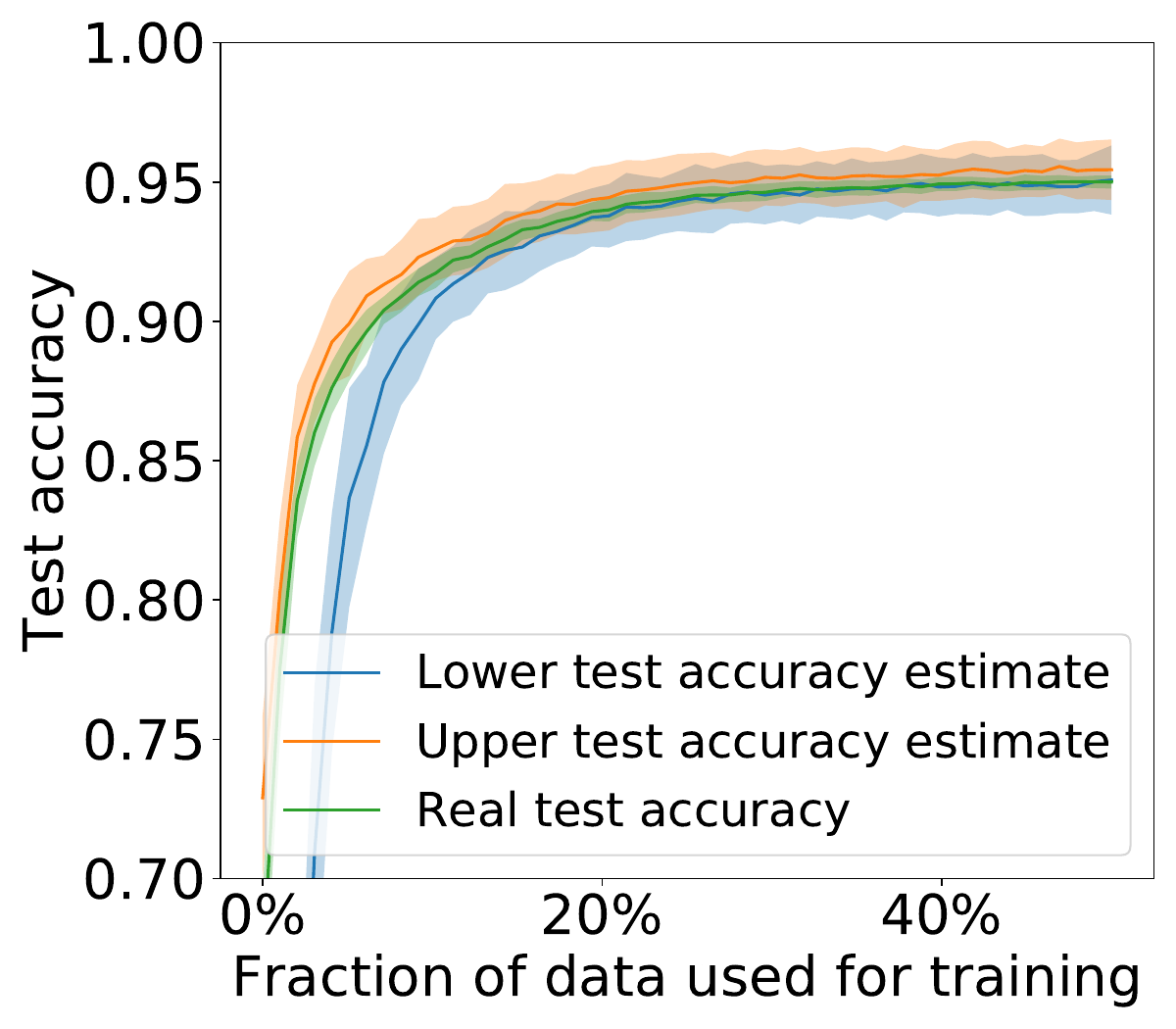}
        \label{fig:estimates_of_the_test_accuracy_495_5_1:mnist}
    }
    \hfill
    \subfigure[Random start size 600, $\mathrm{ANN1}_{100,5}$.]{
        \includegraphics[width=0.22\textwidth]{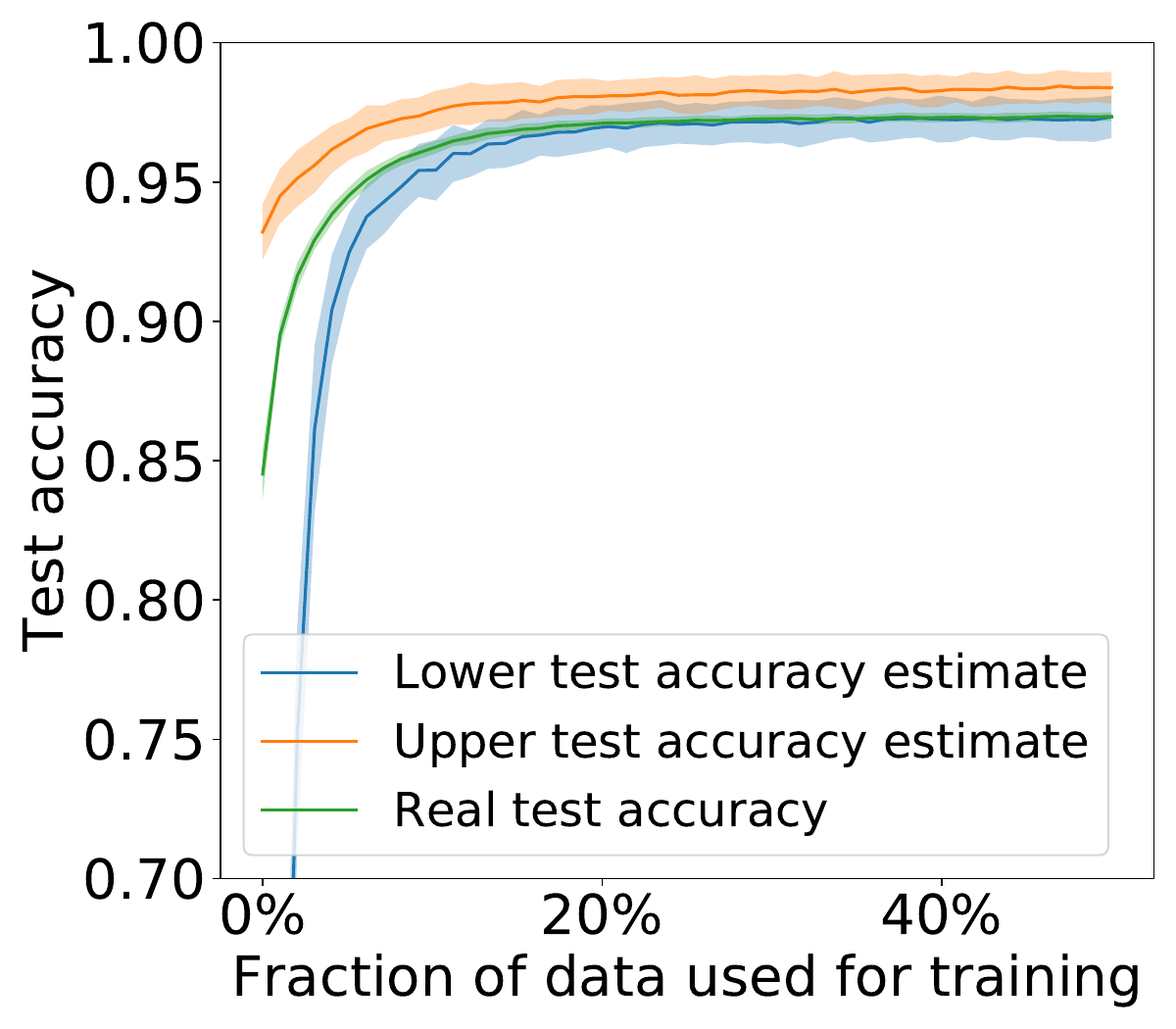}
        \label{fig:estimates_of_the_test_accuracy_495_5_5:mnist}
    }
    \caption{Upper and lower test accuracy estimates for random start of different size and epochs on the MNIST data set, averaged over 50 runs.}
    \label{fig:estimates_of_the_test_accuracy:evaluation:mnist}
\end{figure}

Figure \ref{fig:estimates_of_the_test_accuracy:evaluation:mnist} shows additional experiments using the random start to estimate the test accuracy. The random start sizes  are 600 and 3000, and the types of the networks are  $\mathrm{ANN1}_{100,1}$ and  $\mathrm{ANN1}_{100,5}$. The results show that the upper and lower estimates are better after training on a large fraction on the unlabeled data.  Using a random start of size  600 yields a larger variance compared to a random start of size 3000. Thus a sufficiently large random start is needed in order to make the estimates certain. However, a too large random start might yield a worse lower estimate. 

When using one epoch, as shown in Figures \ref{fig:estimates_of_the_test_accuracy_475_25_1:mnist} and
\ref{fig:estimates_of_the_test_accuracy_495_5_1:mnist}, the upper estimate is close to the real test accuracy. When using 5 epochs, as shown in Figures \ref{fig:estimates_of_the_test_accuracy_475_25_5:mnist} and \ref{fig:estimates_of_the_test_accuracy_495_5_5:mnist}, the upper estimate is considerably higher. This is natural since the more epochs help to learn more about the data. This directly affects the upper estimate, since the data is shared between the test set and the training set in this case. The upper estimate is thus worse if many epochs are used to train the model. Yet, this naive method to estimate the test accuracy might be still useful in practice.

\section{Semi-Supervised Deep Active Learning}

\newcommand{\oxi}{\xi^*}

\paragraph*{Methodology}
\label{sec:method:semi_supervised}

In active learning, it is often assumed that unlabeled data is abundant. So far, in this paper, we have assumed that only the labeled data is used to train a neural network. In this section, we investigate if semi-supervised learning can be used in conjunction with active learning, to benefit from unlabeled data.

For this purpose, we extend the semi-supervised learning method in \cite{pseudo-label-lee-hyun}, where the network assigns \textit{pseudo-labels} to the unlabeled data, and they also contribute to train the network. For each object, the assigned pseudo-label is the most probable class according to the prediction made by the network. This method yields an effective way to use semi-supervised training with deep neural networks. Our extension involves the use of informativeness measures to asses the confidence of the network predictions. In this way, we repeatedly assign those pseudo-labels to unlabeled data that the model is confident about to minimize errors. We call this method the \textit{Gradual Pseudo-Labeling Algorithm} (GPLA).

\paragraph*{The Gradual Pseudo-Labeling Algorithm (GPLA)}
\label{sec:semi_supervised:algorithm}
We perform
GPLA after a network has been trained using some labeled data $L$. For every object $i \in U$, the probability estimates $\hat{y}_{ic}$ are computed for every class $c$ by feeding the data $\mathbf{x}_i$ through the neural network. The probability estimates $\hat{y}_{ic}$ are then used in conjunction with the least confident informativeness measure $I^{LC}_i = -\max_c P_C(y=c | \x_i)\; ,$
where $\xi$ is the \textit{confidence threshold}.We assign to every $i \in S_\xi$ a pseudo-label that matches the predicted class $\argmax_c \hat{y}_{ic}$.
We then update the set of all labeled data by $L \leftarrow L\cup S_\xi$ and the set of all unlabeled data by $U \leftarrow U \setminus S_\xi$
We retrain the network with the newly extended data set $L$ and the process is repeated until only very few objects are added to $S_\xi$. As the network becomes more confident in its predictions, we add more pseudo-labels, hence we call it \textit{gradual} pseudo-labeling. Algorithm \ref{alg:gpla} summarizes the different steps of the procedure.

\begin{algorithm}
\DontPrintSemicolon
  Train network $C$ using all labeled data in $L$.

  Obtain $\hat{y}_{ic}$ by feeding $\mathbf{x}_i$ to $C$, for every $i \in U$.

  Compute $I^{LC}_i$ for every $i \in U$. 

  Construct $S_\xi = \{ i \in U  \mid I_{i}^{LC} < \xi \}$.

  \If{$|S_\xi| > N$}
  {
    Get the pseudo-labels by $\argmax_c \hat{y}_{ic}$ for every $i \in S_\xi$.

    $L \leftarrow L \cup S_\xi$, $U \leftarrow U \setminus S_\xi$.

    Repeat from step 1 until $|S_\xi| \leq N $.
  }

\caption{\label{alg:gpla} The \textit{Gradual Pseudo-Labeling Algorithm} (GPLA), given a confidence threshold $\xi$ and fewest allowed points $N$ in $S_\xi$.}
\end{algorithm}

\paragraph*{Finding a suitable threshold}
\label{sec:semi_supervised:find_good_threshold}

The threshold $\xi$ may have a significant impact on the semi-supervised learning scheme. A too hard threshold, i.e., a $\xi \gtrapprox -1$, makes $S_\xi = \emptyset$, which would be equivalent to only performing supervised active learning. On the other hand, a too soft threshold, i.e., $\xi \lessapprox 0$, causes $S_\xi = U$, which could be suboptimal. We hypothesize that there exists an optimal threshold $\xi^*$ which yields the best performance for GPLA given the current state of the network.
We find a proper $\xi$ by first performing GPLA using several candidate thresholds $\xi \in [-1, 0]$ and measuring the corresponding test accuracy of the network. We perform a $k$-NN ($k$ nearest neighbor) regression as a function of the threshold values and select the $\xi$ that gives the highest accuracy in the regression. Note that  one may use any other suitable regression model instead of $k$-NN regression.

\paragraph*{Experimental setup}

\label{sec:semi_supervised:the_study}
We train an $\mathrm{ANN1}_{n,e}$ network cumulatively on the data selected by a query strategy $A$.
When a certain number $p$ of unlabeled objects are labeled by the oracle through the query strategy, the optimal threshold $\xi^*$ is found for the network by testing 100 uniformly distributed thresholds between -1 and 0. We apply the $k$-NN regression with $k=30$. 
While evaluating the thresholds, we also consider the number of erroneously labeled objects to investigate if the number of errors the algorithm makes changes depending on $\xi$.  Given $\oxi$, we then employ GPLA to obtain the new test accuracy using $N = 150$. Training of GPLA is done in the cumulative mode.
The number of examined labeled images $p$ is equal to $500$, $1000$, $2500$, $5000$, $10000$, $20000$, and $50000$, where there are $60000$ images in total. We choose the query strategy $A$ to be either $M$ or $R$.
We use an $\mathrm{ANN1}_{100,1}$ network for the MNIST data set to evaluate the usefulness of GPLA. 

\paragraph*{Results and discussion}
\label{sec:results:semi_supervised_training}

\begin{figure}[t]
    \centering
    \hspace{-4mm}
    \subfigure[
    Accuracy and $k$-NN regression.
    ]{\includegraphics[width=0.28\textwidth]{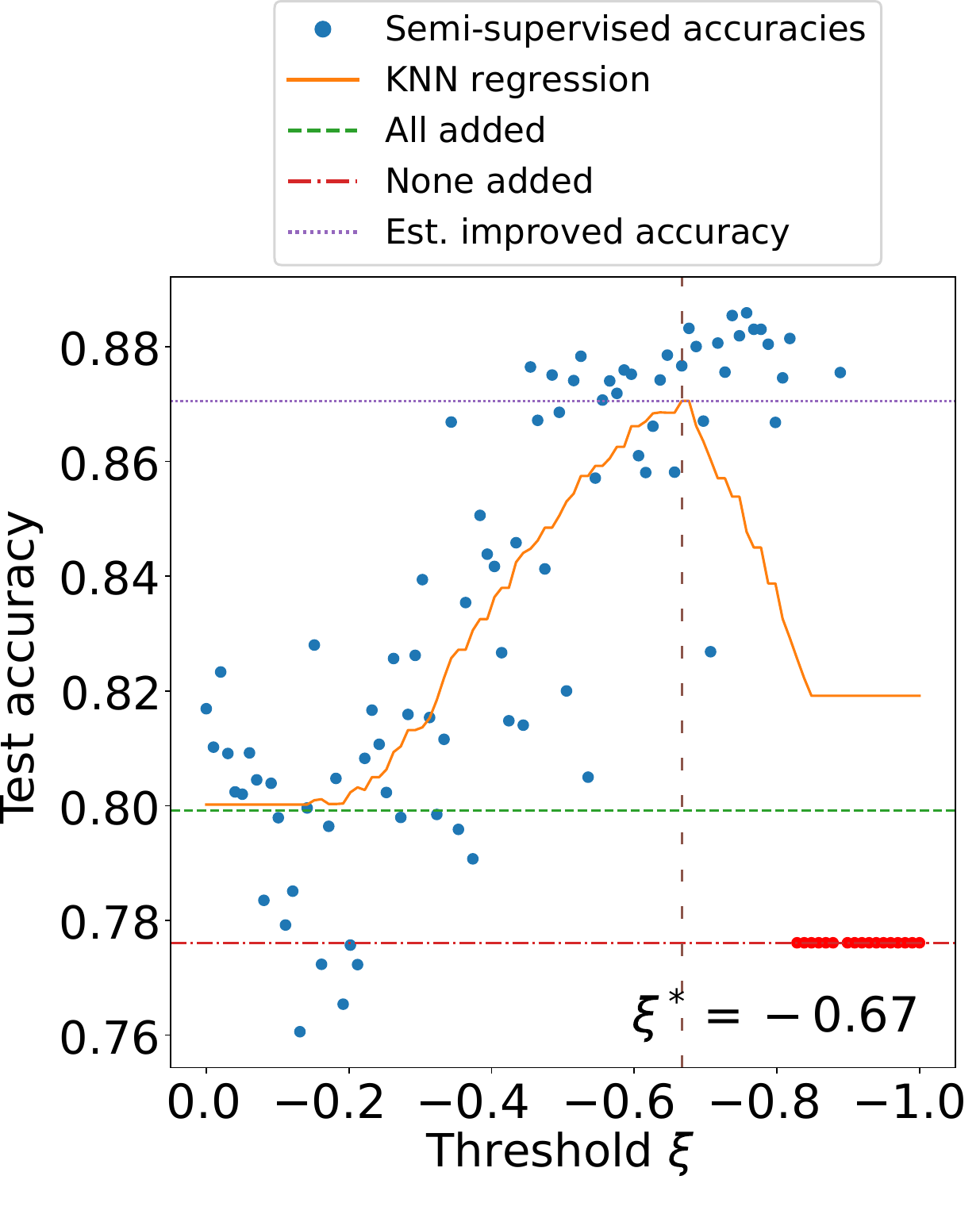} \label{fig:semi_supervised_training:KNN-regression:margin:mnist:a}}
    \hspace{-2mm}
    \subfigure[
    Wrong  pseudo-labels.
    ]{\includegraphics[width=0.28\textwidth]{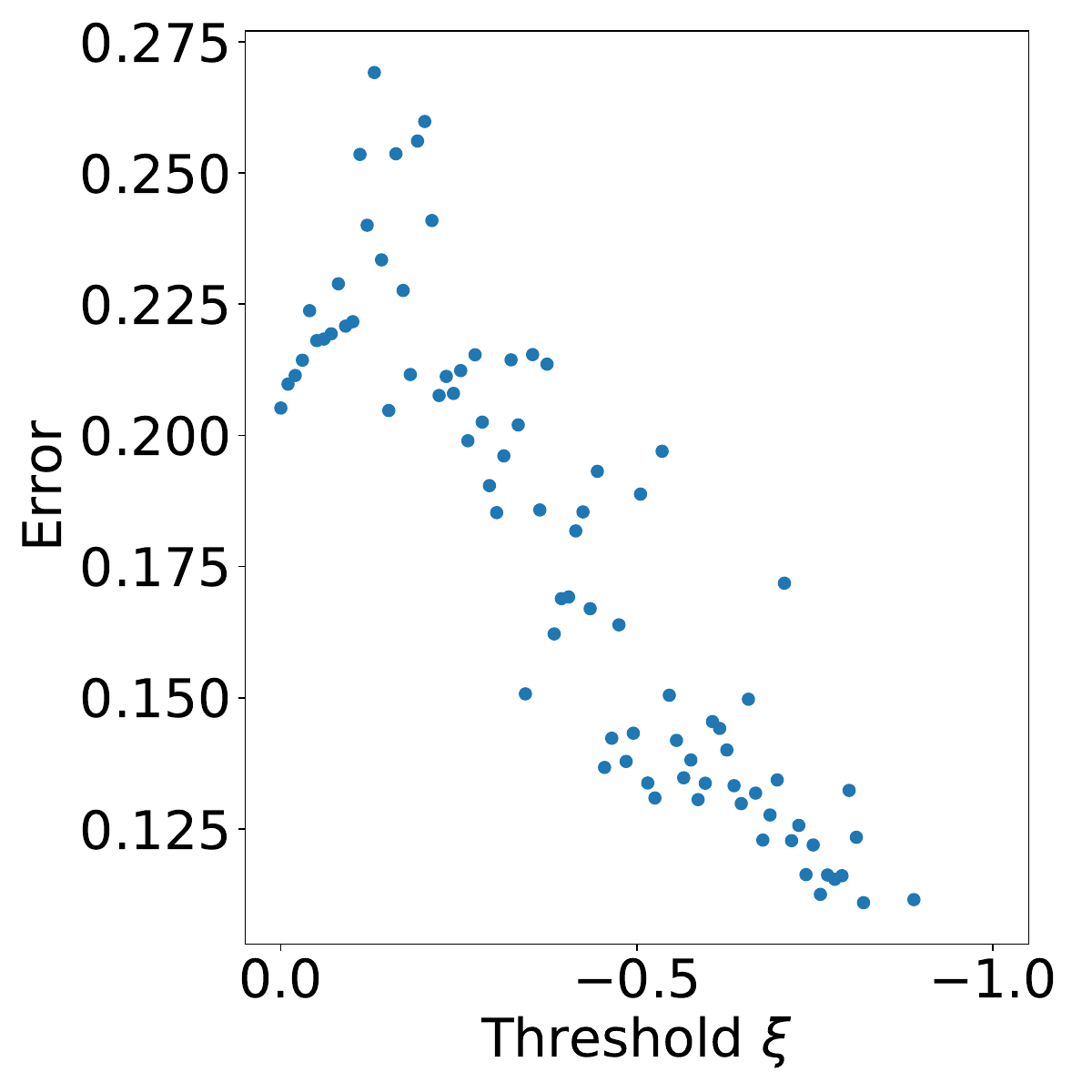} \label{fig:semi_supervised_training:margin:mnist:error}}
    \caption{Test accuracy and pseudo-label assignment error with GPLA.
    }

    \label{fig:semi_supervised_training:margin:mnist}
\end{figure}
\begin{figure}[t]
    \centering
    \hspace{-4mm}
    \subfigure[$\mathrm{ANN1}_{100,1}$ network.]{\includegraphics[width=0.24\textwidth]{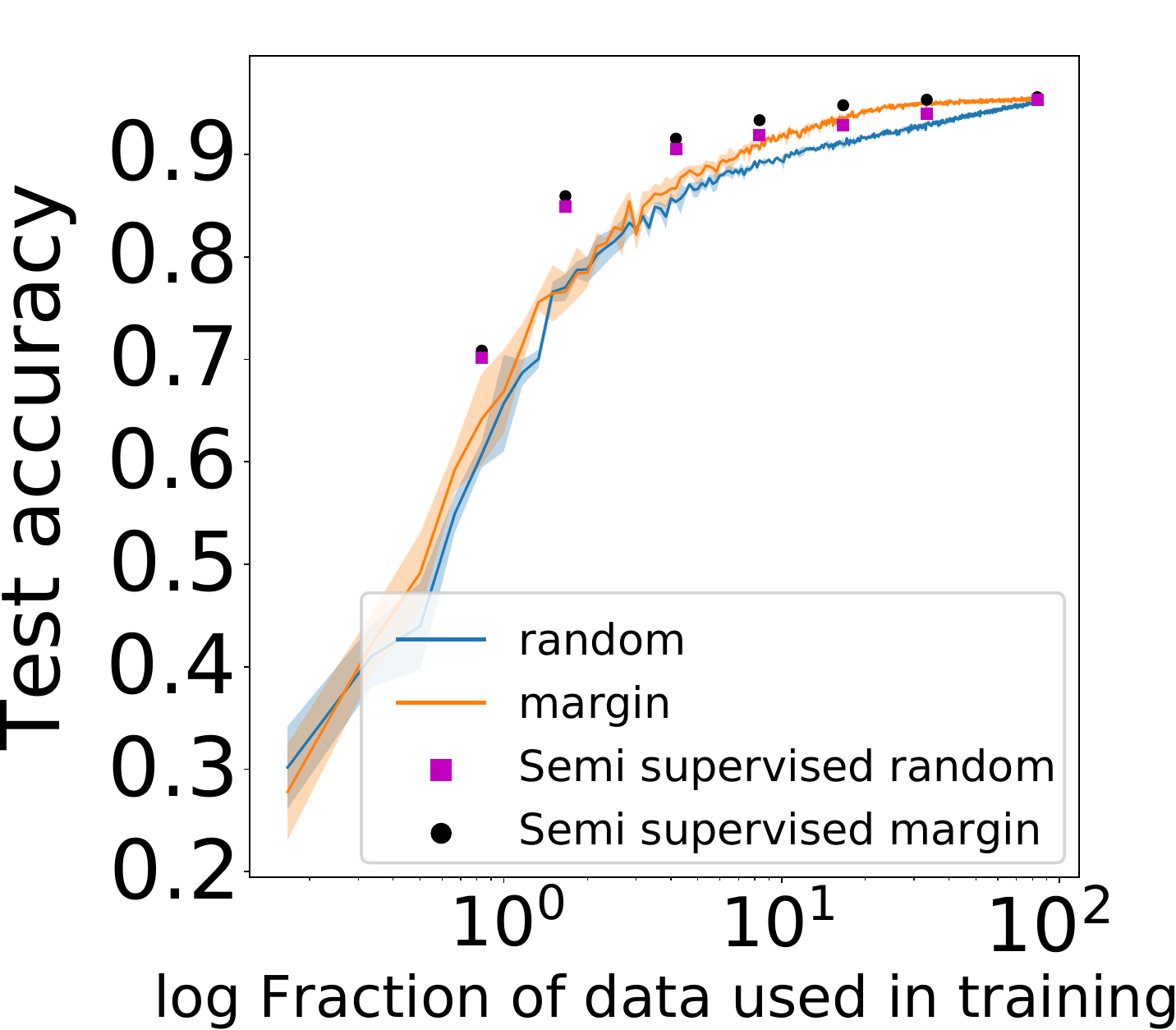} \label{fig:semi_supervised_training:improved:mnist:1epoch}}
    \hspace{-2mm}
    \subfigure[$\mathrm{ANN1}_{1000,5}$ network.]{\includegraphics[width=0.24\textwidth]{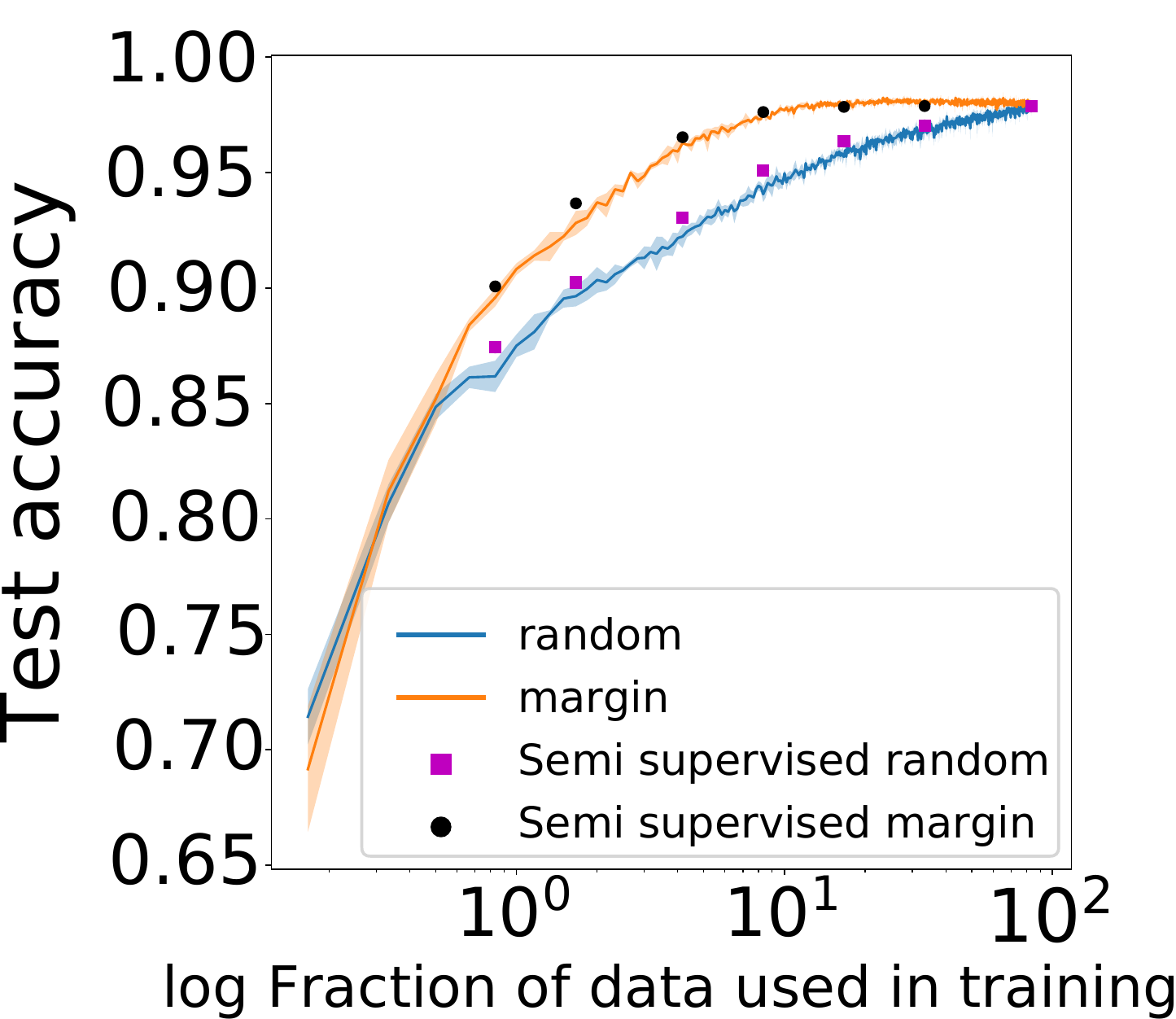} \label{fig:semi_supervised_training:improved:mnist:5epoch}}
    \caption{Test accuracy before and after semi-supervised active learning.
    The optimal semi-supervised accuracies $a_{\xi^*}$ are shown as dots.
    }
    \label{fig:semi_supervised_training:improved:fashion:mnist}
\end{figure}

Figure \ref{fig:semi_supervised_training:KNN-regression:margin:mnist:a}
shows the results w.r.t. the threshold $\xi$ for $p = 1000$, where we observe improvement in the test accuracy with semi-supervised learning for the 100 candidate thresholds $\xi \in [-1, 0]$. The network is $\mathrm{ANN1}_{100,1}$ trained with the margin query strategy. 
As the thresholds get harder (that is, closer to -1) we see an improvement in test accuracy, i.e., an almost (negative) correlation between the threshold and the test accuracy.
The result indicated by \textit{`All added'} is the average test accuracy for all $\xi$ for which $S_\xi = U$, i.e., the runs wherein all the objects are assigned a pseudo-label at the first iteration. Thus it corresponds to the way the pseudo-labels are assigned in \cite{pseudo-label-lee-hyun}.
Figure \ref{fig:semi_supervised_training:margin:mnist:error} shows the fraction of erroneously assigned pseudo-labels for different thresholds. Consistently, we observe an almost (positive)  correlation between the threshold and the fraction of erroneously assigned pseudo-labels (except for $\xi$ close to zero). This supports our hypothesis that introducing a threshold may reduce the number of erroneously assigned pseudo-labels. The choice of $\xi = -0.8$ almost halves the number of errors, compared to the no threshold case ($\xi = 0$).

Figure \ref{fig:semi_supervised_training:margin:mnist} suggests that the optimal threshold $\xi^*$ is the hardest (smallest) threshold that still allows the network to assign pseudo-labels, i.e., a threshold for which $S_\xi\neq \emptyset$. Thus, in principle, one may avoid using $k$-NN regression, and a validation set, to select a proper $\xi$.
Such a choice makes sense since a hard threshold suggests choosing only very confident pseudo-labels. Then, by repeating the procedure possibly new confident pseudo-labels are obtained. Therefore, at the end, we compute all the confident pseudo-labels obtained through either the true labels or the other confident pseudo-labels, and together with true labels, we use them for training the network.
The test accuracy in Figure \ref{fig:semi_supervised_training:KNN-regression:margin:mnist:a} increases from 78\% to 88\%, after applying the gradual pseudo-labeling assignment with the hardest $\xi$, when 1000 images from in total 60000 images are labeled by the oracle.

Figure \ref{fig:semi_supervised_training:improved:fashion:mnist} shows the test accuracy before and after pseudo-labeling is applied. We observe that the semi-supervised test accuracy is higher regardless of whether we use $M$ or $R$ for query strategy.
The relative increase in accuracy is larger for smaller $p$'s. It is natural that the accuracy for  a large $p$ can not be improved much, since the network is almost as good as it can be given correct labels. Therefore, the proposed semi-supervised learning scheme can be used in combination with different query strategies, to be applied after every active learning step to attain the full potential of the unlabeled data set and improve the test accuracy. The improvement in test accuracy can be significant.
We note that our approach even potentially allows us to perform semi-supervised learning during the active learning step, making the model more certain about which data to query next.

\begin{figure}[h!]
    \centering
    \includegraphics[width = 0.298\textwidth]{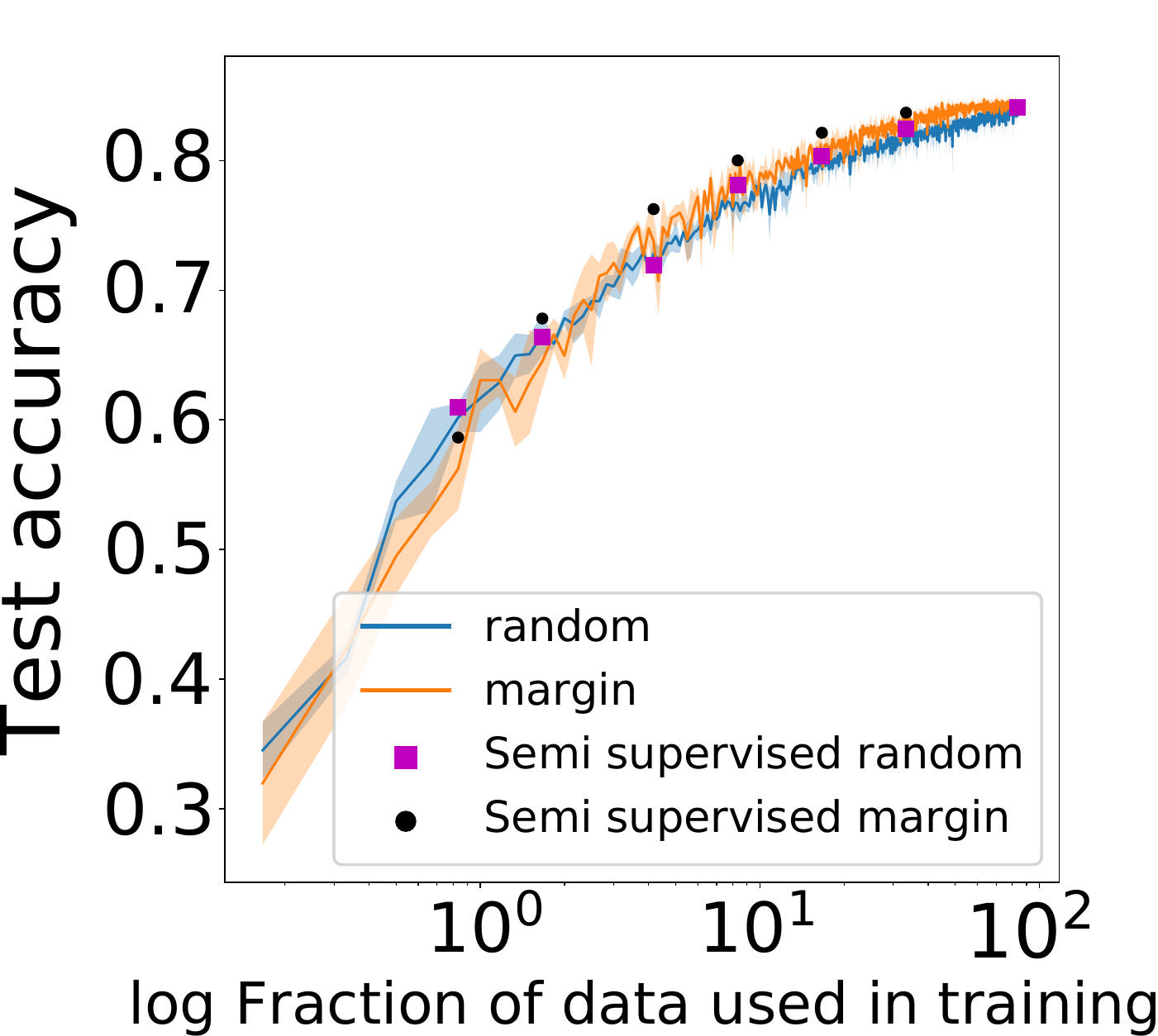}
    \caption{Test accuracy before and after semi-supervised active learning.} 
    \label{fig:app:semi_supervised_acc:fashion}
\end{figure}

Finally, Figure \ref{fig:app:semi_supervised_acc:fashion} shows the  results for the Fashion-MNIST data set. The results demonstrate improvement, although it is not as significant as on the MNIST data set.

\section{Conclusion}
We studied in a consistent way several data-centric and model-centric aspects of active learning for neural network models. Our studies provide further insights on this task and help us for a  better understanding of the concerns that might arise when using active learning to train a neural network model, e.g., training mode, query strategies, availability of unlabeled data, initial training, and network specifications. We finally investigated semi-supervised active learning via propagating and using pseudo-labels.

\bibliography{bibliography}
\bibliographystyle{plain}

\end{document}